\documentclass[10pt,twocolumn,letterpaper]{article}
 
\pdfoutput=1 
\usepackage{iccv}
\usepackage{times}
\usepackage{epsfig}
\usepackage{graphicx}
\usepackage{amsmath}
\usepackage{amssymb}
\usepackage{subfigure}
 
\usepackage[usenames, dvipsnames]{color}

\usepackage[pagebackref=true,breaklinks=true,letterpaper=true,colorlinks,bookmarks=false]{hyperref}
 
\iccvfinalcopy %
 
\def\iccvPaperID{1082} %

\ificcvfinal\fi
\begin{document}

\title{Back to RGB: 3D tracking of hands and hand-object interactions\\ based on short-baseline stereo}

\author{Paschalis Panteleris\\
Institute of Computer Science, FORTH\\
{\tt\small padeler@ics.forth.gr}
\and
Antonis Argyros\\
Computer Science Department, University of Crete and Institute of Computer Science, FORTH\\
{\tt\small  argyros@ics.forth.gr
}
}
 
\maketitle
\begin{abstract}
We present a novel solution to the problem of 3D tracking of the articulated motion of human hand(s), possibly in interaction with other objects. 
The vast majority of contemporary relevant work capitalizes on depth information provided by RGBD cameras. In this work, we show that accurate and efficient 3D hand tracking is possible, even for the case of RGB stereo. A straightforward approach for solving the problem based on such input would be to first recover depth and then apply a state of the art depth-based 3D hand tracking method. Unfortunately, this does not work well in practice because the stereo-based, dense 3D reconstruction of hands is far less accurate than the one obtained by RGBD cameras. Our approach bypasses 3D reconstruction and follows a completely different route: 3D hand tracking is formulated as an optimization problem whose solution is the hand configuration that maximizes the color consistency between the two views of the hand. We demonstrate the applicability of our method for real time tracking of a single hand, of a hand manipulating an object and of two interacting hands. The method has been evaluated quantitatively on standard datasets and in comparison to relevant, state of the art RGBD-based approaches. The obtained results demonstrate that the proposed stereo-based method performs equally well to its RGBD-based competitors, and in some cases, it even outperforms them.
\end{abstract}
 
\section{Introduction}
\label{sec:intro}
 
\begin{figure}[t]
\begin{center}

\includegraphics[width=0.99\columnwidth]{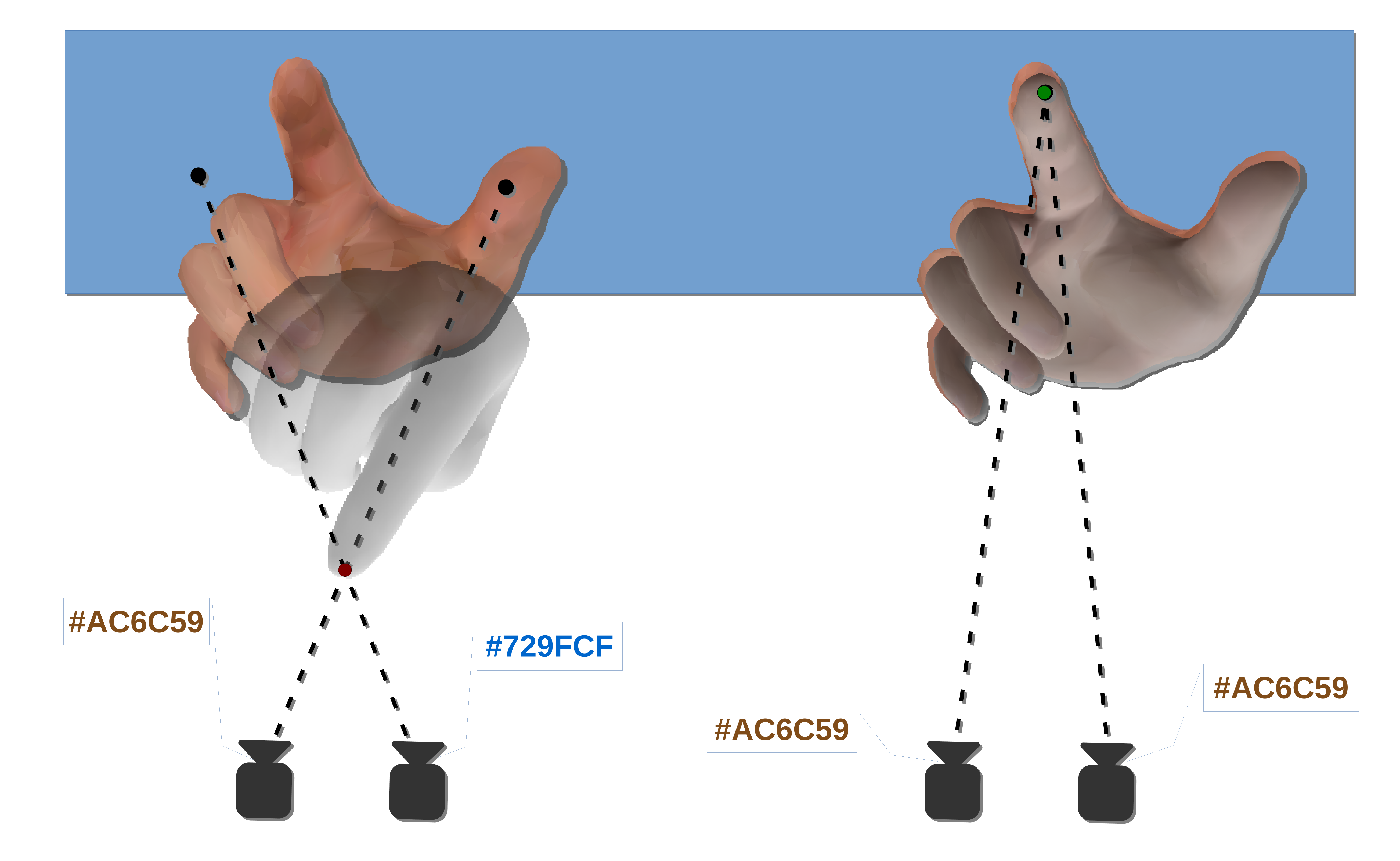} 
\caption{\label{fig:concept} Illustration of the key idea in this work. Left: an actual hand (in skin color) and a wrong hypothesis about its 3D pose and articulation (gray color). The wrong hypothesis about the 3D configuration of the hand leads to pixels with dissimilar colors in the stereo camera views. Right: the hand in the same pose and a correct hand hypothesis which back-projects to pixels with identical colors on the stereo camera views. Hand tracking is formulated as an optimization problem that seeks the hand configuration that maximizes the color consistency of the two views of the hand.}
 \end{center}
\end{figure}
 
The vision-based recovery of the 3D pose of human hands is an interesting and important problem in computer vision. Humans use their hands all the time and in several ways either to interact with the physical world or to communicate with other humans. An accurate, robust and real time solution to the problem has a huge impact in a number of application domains including HCI, HRI, medical rehabilitation, sign language recognition, etc. %
 
\definecolor{hybrid}{rgb}{0.0, 0,0}
\definecolor{model}{rgb}{0,0,0}
\definecolor{discr}{rgb}{0,0,0.0}

\newcommand{\sdepth}{
\color{model}
\cite{oikonomidis2011efficient} \cite{schmidt2014dart} 
\cite{bray2007smart} \cite{kim2012digits} 
\cite{MakrisKyriazisArgyros2015a} 
\cite{melax2013dynamics} \cite{taylor2014user} \cite{khamis2015learning} 
\cite{MakrisArgyros2015a} \cite{zhao2012combining}
\cite{fleishman2015icpik}

\color{discr}
\cite{tang2015opening} \cite{keskin2012hand} \cite{oberweger2015training}
\cite{xu2013efficient} \cite{sun2015cascaded} \cite{tang2013real} \cite{li20153d}

\color{hybrid}
\cite{taylor2016efficient} \cite{tagliasacchi2015robust} 
\cite{sridhar2015fast} \cite{sharp2015accurate} \cite{tompson2014real}
\cite{qian2014realtime} \cite{poier2015hybrid}
}

\newcommand{\odepth}{
\color{model}\cite{panteleris20153d} \cite{kyriazis2014scalable} \cite{hamer2010object}
\cite{kyriazis2013physically} \cite{hamer2009tracking}
\newline
\color{discr}\cite{rogez2015first} \cite{rogez2015understanding}
\newline
\color{hybrid}
\cite{sridhar2016real}
}

\newcommand{\tdepth}{
\color{model}\cite{oikonomidis2012tracking} \cite{OikonomidisLourakisArgyros2014a}
\color{discr}
\newline $\times$
\newline
\color{hybrid}\cite{sridhar2013interactive}
}

\newcommand{\smulti}{
\color{model}\cite{oikonomidis2010markerless} \cite{sridhar2014real}
\cite{OikonomidisKyriazisTzevanidisEtAl2013a}
\newline 
\newline
\color{discr}
\cite{de2006regression} \newline
\color{hybrid} $\times$
}

\newcommand{\omulti}{
\color{model}\cite{OikonomidisKyriazisArgyros2011a} \cite{wang2013video}
\newline
\color{discr} $\times$
\newline 
\color{hybrid}$\times$
}

\newcommand{\tmulti}{
\color{model} $\times$ \newline
\color{discr} $\times$ \newline
\color{hybrid} \cite{ballan2012motion} \cite{tzionas2015capturing}
}

\newcommand{\sstereo}{
\color{model}\cite{rehg1994visual} %
\newline
{\bf [Proposed]}
\newline
\color{discr}\cite{rosales20013d} \newline
\color{hybrid} $\times$
}

\newcommand{\ostereo}{
\color{model}{\bf [Proposed]} \newline
\color{discr} $\times$ \newline
\color{hybrid} $\times$
}

\newcommand{\tstereo}{
\color{model}{\bf [Proposed]}\newline
\color{discr} $\times$ \newline
\color{hybrid} $\times$
}

\newcommand{\smono}{
\color{model}\cite{de2011model} %
\cite{maccormick2000partitioned} %
\cite{stenger2001model} %
\cite{wu2001capturing} %
\cite{sudderth2004visual} %
\cite{wang2009real} %
\cite{thayananthan2003shape} %
\cite{heap1996towards}

\color{discr}
\cite{athitsos2003estimating} %
\cite{wu2000view}
\cite{romero2009monocular}
\color{hybrid} 
\newline $\times$
}

\newcommand{\omono}{
\color{model} $\times$ \newline
\color{discr}
\cite{romero2010hands} \newline
\color{hybrid} $\times$
}

\newcommand{\tmono}{
\color{model} $\times$ \newline
\color{discr} $\times$ \newline
\color{hybrid} $\times$
}

\newcommand{\rtypeA}{
\color{model}G
\newline
\newline
\color{discr}D
\newline
\color{hybrid}H
}

\newcommand{\rtypeB}{
\color{model}G
\newline
\color{discr}D
\newline
\color{hybrid}H
}

\newcommand{\rtypeC}{
\color{model}G
\newline
\color{discr}D
\newline
\color{hybrid}H
}

\begin{table*}[t]
\footnotesize
\begin{center}
\begin{tabular}{|p{0.1\textwidth}|p{0.04\textwidth}|p{0.27\textwidth}|p{0.11\textwidth}|p{0.12\textwidth}|p{0.15\textwidth}|} \hline
{\bf Problem}     & \bf{Type}    & \bf{RGB-D} & \bf{Multi RGB} & \bf{Stereo RGB} & \bf{Mono RGB} \\  \hline\hline 
Single hand       & \rtypeA      & \sdepth    & \smulti          & \sstereo    & \smono \\ \hline\hline
Hand-Object       & \rtypeB      & \odepth    & \omulti          & \ostereo    & \omono \\ \hline\hline
Two hands         & \rtypeC      & \tdepth    & \tmulti          & \tstereo    & \tmono \\ \hline\hline
\end{tabular}
\end{center}
\caption{An overview of related work and the positioning of the proposed approach in it. Rows: the variants of the basic problem (in increasing problem dimensionality and complexity). Columns: the type of required input (in decreasing wealth of information content). For each problem variant, we categorize generative methods (G), discriminative (D), and hybrid ones (H).}
\label{tab:review}
\end{table*}

The problem of 3D hand tracking is challenging. Difficulties arise due to the highly articulated structure of the hand which results in ambiguous poses and self-occlusions. These problems escalate when hands interact with each other and/or manipulate objects. On top of these intrinsic difficulties, some application areas impose constraints on the spatial and temporal resolution of the images that feed a 3D hand pose estimation/tracking algorithm.
 
During the last couple of years, a number of methods for 3D hand tracking or single-frame 3D hand pose estimation have appeared. Most of the methods address the problem under the assumption that a single acting hand is observed. To a lesser extend, solutions for the problem of tracking hand-object interactions have also been proposed. The common characteristic of contemporary approaches is that they rely on RGBD or depth data. Dense and accurate depth information proves sufficient for arriving at accurate pose estimation. Still, it would be highly desirable if accurate and efficient 3D hand and hand/object tracking could be achieved on the basis of RGB information, alone. This would make 3D hand tracking possible by the vast majority of today's camera systems that exist everywhere (including smartphones, tablets, etc) and do not record depth information. It would also make 3D hand tracking possible in outdoor environments where several active RGBD sensors do not provide reliable information.
 
In this work we address exactly this challenge. We present the first method that performs detailed, accurate and real time 3D tracking of scenes comprising  hands based on a conventional, passive, short-baseline, RGB stereo. The naive approach to this problem would be to first perform 3D reconstruction and then employ a state of the art approach for depth-based hand tracking/pose estimation. However, as we show experimentally, this approach does not produce reliable results. Depth information is either noisy, or sparse or smoothed-out to support accurate 3D tracking. Thus we follow a completely different path. We capitalize on the very successful hypothesize-and-test, generative tracking paradigm. We assume a 3D model of the object(s) to be tracked (i.e., a hand, a hand-object constellation, two hands). As shown in Figure~\ref{fig:concept}, a hypothesis regarding the configuration of such a model gives rise to hypotheses on the 3D structure of the scene. A correct model hypothesis leads to photo-consistent views. Tracking the model is then formulated as an optimization problem that seeks the model configuration that maximizes stereo color consistency.

The proposed method has been evaluated extensively in standard datasets and in comparison to existing, state of the art methods that rely on depth information. We investigate three interesting problem subclasses, namely (a) single hand tracking, (b) a hand interacting with a rigid object and (c)~two interacting hands. The obtained results lead to the conclusion that the proposed stereo-based algorithm can be as good and can even outperform their RGBD-based counterparts, both in terms of accuracy and speed.

\begin{figure*}[t]
\minipage{0.12\textwidth}
\includegraphics[height=0.16\textheight]{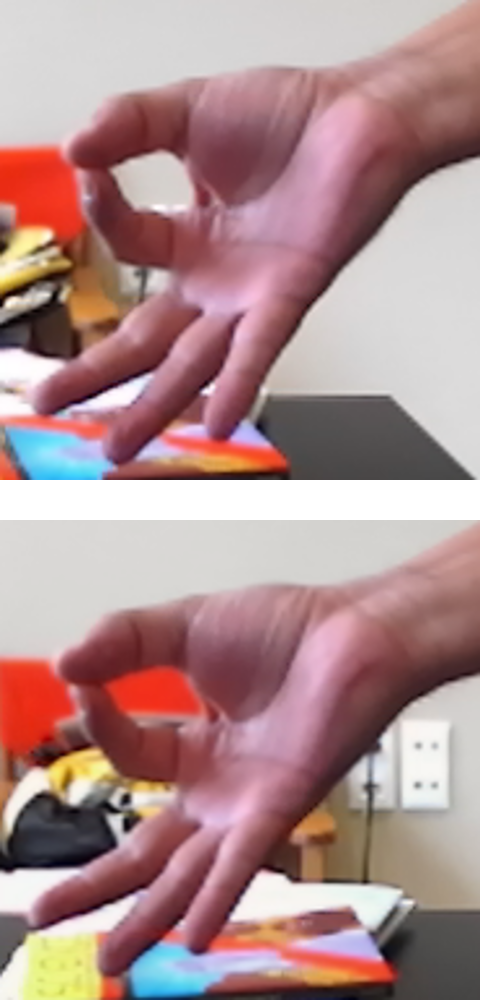}
\endminipage
\minipage{0.12\textwidth}
\includegraphics[height=0.16\textheight]{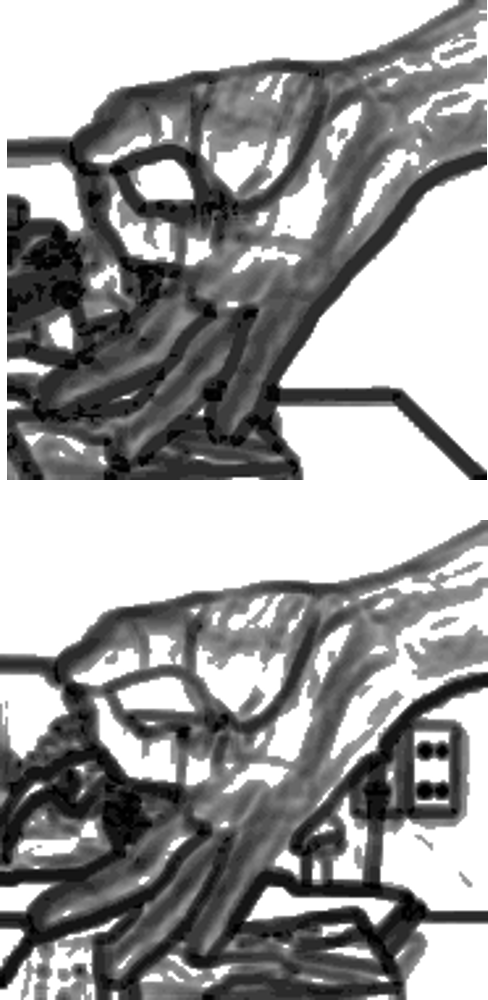}
\endminipage
\minipage{0.12\textwidth}
\includegraphics[height=0.10\textheight]{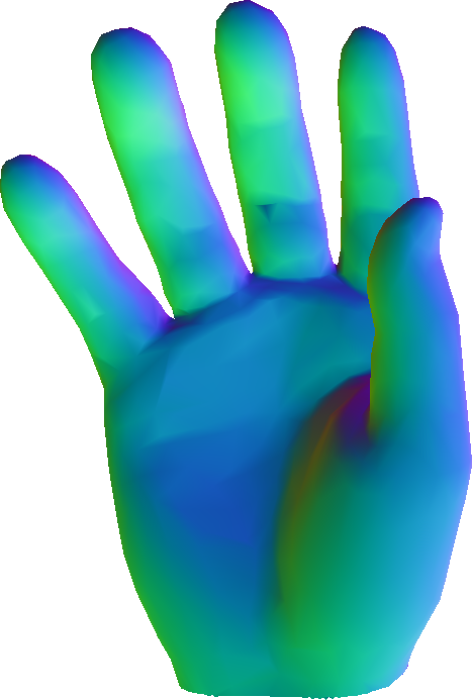}
\endminipage
\minipage{0.22\textwidth}
\includegraphics[height=0.16\textheight]{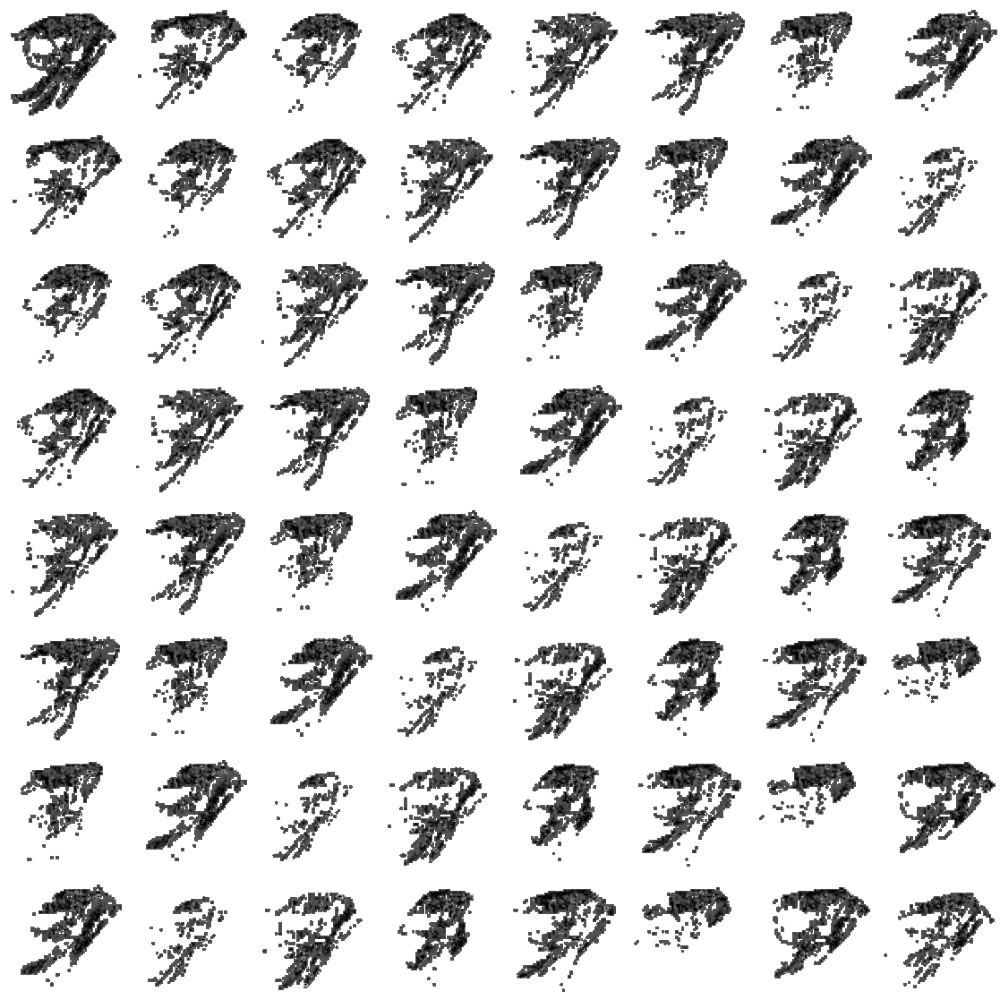}
\endminipage
\minipage{0.16\textwidth}
\includegraphics[height=0.10\textheight]{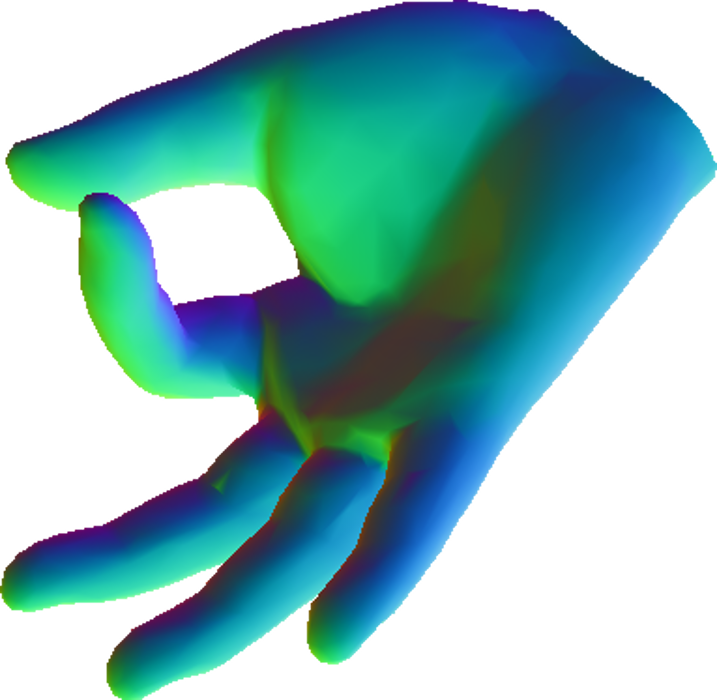}
\endminipage
\minipage{0.16\textwidth}
\includegraphics[height=0.16\textheight]{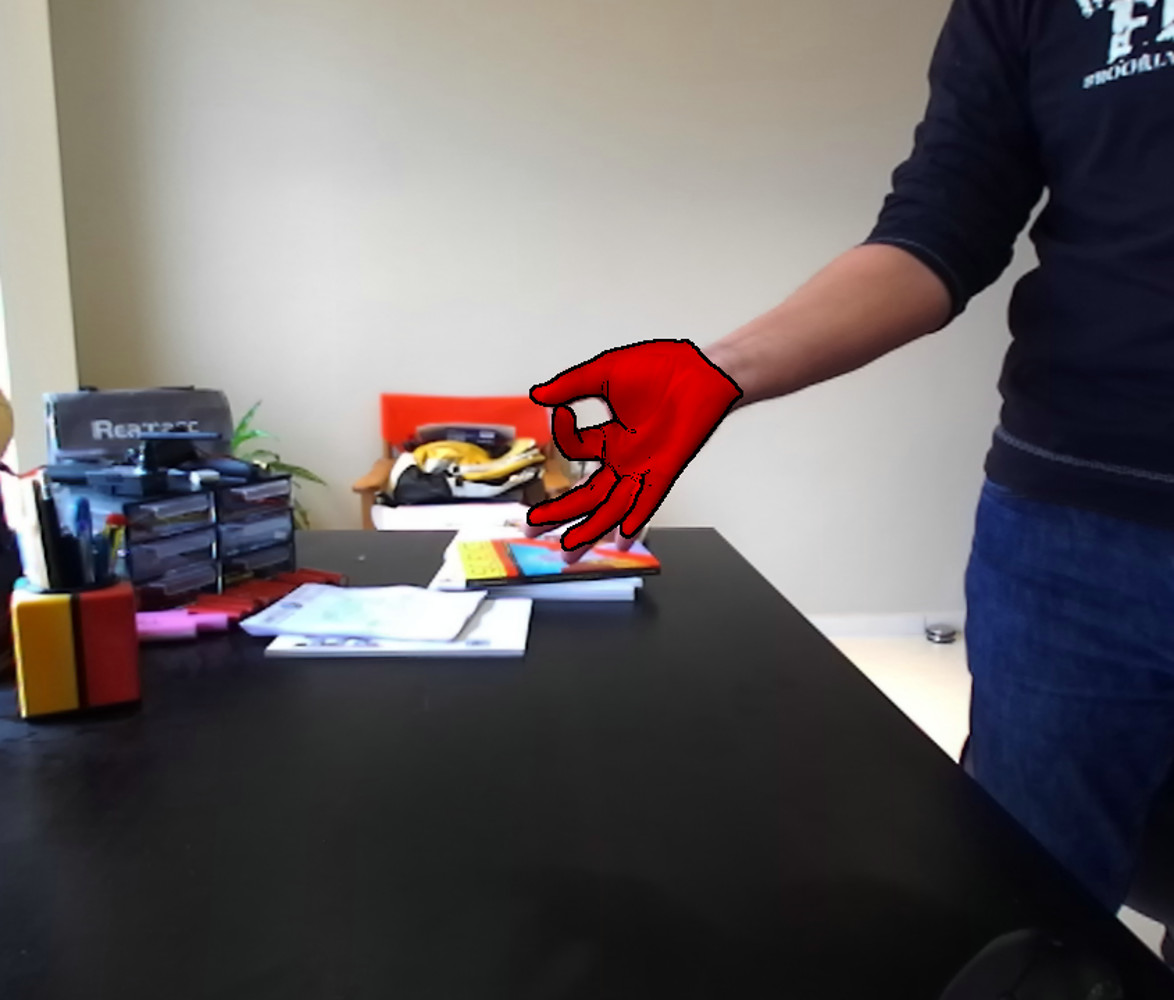}
\endminipage
\vspace*{0.2cm}
\minipage{0.12\textwidth}
\centering (a)
\endminipage
\minipage{0.12\textwidth}
\centering (b)
\endminipage
\minipage{0.12\textwidth}
\centering (c)
\endminipage
\minipage{0.22\textwidth}
\centering (d)
\endminipage
\minipage{0.16\textwidth}
\centering (e)
\endminipage
\minipage{0.24\textwidth}
\centering (f)
\endminipage
\caption{\label{fig:method} An overview of the proposed pipeline. (a) The stereo pair input (b) the computed distinctiveness maps (cropped around the last hand position) (c) the skinned mesh of the employed hand model (d) the color consistency scores after the evaluation of 64 PSO particles for the last generation of the optimization process (the top scoring hypothesis is the top left), (e) the hand model configured in the highest scoring pose and, (f) the solution in (e) rendered and superimposed on the left image of the stereo pair.}
\end{figure*}

\section{Related work}
\label{sec:relatedwork}
The three tracking scenarios correspond to problems of increasing dimensionality/complexity. Existing solutions can be classified along two important dimensions. 
 
\noindent\textbf{Discriminative vs generative vs hybrid:} Discriminative methods learn a mapping from observations to poses. Generative ones fit a model to the set of available observations. Discriminative methods are faster, less accurate and do not require initialization, i.e., perform single frame pose estimation. Generative ones are more accurate, require initialization and perform tracking  to exploit temporal continuity. Hybrid methods try to couple the benefits from both worlds by employing a discriminative component to arrive at a coarse solution which is then refined by a generative, model-based component. 
 
\noindent\textbf{Based on their input:} There are methods that rely on RGBD sensors, multicamera setups, stereo RGB cameras or single camera input.
Table~\ref{tab:review} provides an overview of the research in the field, which leads to a number of interesting conclusions: (a) The vast majority of solutions are based on input from depth sensors. (b) There are very few stereo-based methods, all of which deal with single hand tracking. (c) For the problems of hand-object and two-hands tracking, there is no available stereo method. Moreover, although it appears that there are several methods for monocular RGB tracking of a single hand, these methods only produce limited information regarding hand pose and in very constrained settings.  
 
{\noindent \bf Our contribution:} 
This paper proposes a generative, model-based tracking framework that~(a) applies to all three problem instances and~(b) uses a compact, short-baseline, calibrated stereo pair. As such, it covers a significant gap in the existing literature. As shown in Section~\ref{sec:results}, the proposed framework constitutes the first practical approach to the problem(s) based on such input and manages to provide solutions that are as good in terms of accuracy and speed as those obtained by their RGBD/depth-based competitors.
 
\section{The proposed method}
\label{sec:method}
Figure~\ref{fig:method} illustrates the proposed pipeline. Although this is shown for the problem of single hand tracking, it is straightforward to extend it to cover more complicated tracking scenarios such as tracking a hand in interaction with an object or two interacting hands. 

We assume images acquired by a calibrated stereo pair (Figure~\ref{fig:method}a). We also assume a 3D hand model (skinned 3D mesh, Figure~\ref{fig:method}c) that can be articulated in 3D space based on its own intrinsic (kinematic model, articulation) and extrinsic (absolute 3D position and orientation) parameters (Section~\ref{sec:model}). A given hypothesis about the hand configuration provides a hypothesis about the 3D location of every point of the hand model. Figure~\ref{fig:concept} illustrates this for the cases of a wrong (left) and a correct (right) hypothesis. If the model hypothesis is wrong, then, different physical points are projected in the two stereo views. Therefore, it is quite likely that the corresponding 2D image points will have inconsistent appearance (e.g., different color). In contrast, if the model hypothesis is correct (Figure~\ref{fig:concept}, right), the color consistency of all pairs of projections of the hand model points on the two stereo images is maximized. Provided that a reasonable measure of color consistency can be defined, tracking the model can be formulated as an optimization problem that seeks the hand pose that maximizes this color consistency (Figure~\ref{fig:method}d-f). In Section~\ref{sec:objective} we do provide such a quantification of color consistency and we employ Particle Swarm Optimization (PSO)~\cite{Clerc2002} in order to maximize it (Section~\ref{sec:pso}). 
The defined objective function measures color consistency by measuring color similarity, weighted by the distinctiveness of the corresponding points. The intuition behind this choice is that color consistency over uniformly colored areas should weight less than color consistency of distinctive points. 
Our measure of distinctiveness (Section~\ref{sec:conf}) is based on the analysis of the Harris corner detector~\cite{harris1988combined}. Sample distinctiveness maps for the stereo pair of images of Figure~\ref{fig:method}a are shown in Figure~\ref{fig:method}b.
 
\subsection{Observing and modelling the scene}
\label{sec:model}
Observations come from a calibrated stereo pair of RGB cameras. The intrinsic (focal length, distortion, camera center) and extrinsic (relative position, orientation) parameters for each camera are computed using standard camera calibration techniques~\cite{bouguet2004camera}. The images of each  stereo pair are temporally synchronized and undistorted before further processing.

In this work, we consider the tracking of hands and rigid objects. For modelling hands we employ the anatomically consistent and visually realistic hand model provided by {\sl libhand}~\cite{libhand} (Figure~\ref{fig:method}c). This is a skinned model of a right hand consisting of $22$ bones. In order to be directly comparable with results of existing methods~\cite{oikonomidis2011efficient,oikonomidis2012tracking}, the original model was adapted by removing the wrist vertices and root bone and by allowing mobility with $26$ degrees of freedom. 
The adapted hand model consists of $21$ bones and of a 3D mesh with $1491$ vertices. The configuration of each hand is represented by $27$ parameters: Three for the hand position, four for the quaternion representation of the hand rotation and four articulation angles for each of the five fingers. 
The model of the left hand is a mirrored version of the model for the right hand. Using the $27$ hand model parameters and the camera parameters (intrinsics + extrinsics) we can render any configuration of the hand model on the stereo pair.  
 
{\sl Libhand} provides a realistic texture for the hand model. This information is not used during the tracking. %
However, the textured hand was used in order to render the synthetic datasets used for the quantitative evaluation of the method, as it will be detailed in Section~\ref{sec:results}. 
 
Rigid objects in the scene have $6$ degrees of freedom, modelled with $7$ parameters, $3$ for the object position and $4$ for the quaternion representation of their rotation. An observed scene can thus be represented as a multi-dimensional vector with as many dimensions as the sum of the numbers of parameters of the individual objects in it.
 
\subsection{Distinctiveness maps}
\label{sec:conf}
When checking the color consistency between image points, it is preferable to give more emphasis to the similarity of distinctive ones. This is because the similarity of distinctive points bears more information content than the similarity of points in uniformly colored regions. 
Thus, we create two distinctiveness maps, one for each of the input images. Our approach for defining distinctiveness borrows from related work in the field of corner detection~\cite{harris1988combined}.
More specifically, for each pixel $p$ in an image we compute the principal curvatures $\lambda_1$ and $\lambda_2$ of the local auto-correlation  function in a neighborhood of $B \times B$ pixels centered on $p$. Without loss of generality, we assume that $\lambda_1 \geq \lambda_2$. We use $B=3$ in order to preserve fine image structures. 
Small $\lambda_1$ and $\lambda_2$ values mean a uniformly colored region. Large $\lambda_1$ and $\lambda_2$ values indicate a corner. Finally, if $\lambda_1$ is significantly larger that $\lambda_2$, the area around $p$ is an edge. 
 
For a definition of distinctiveness we could use the Harris corner detector~\cite{harris1988combined} response function:
\begin{equation}
c_h = \lambda_1 \cdot \lambda_2 - k \cdot (\lambda_1 + \lambda_2)^2.
\end{equation}
This option has been evaluated experimentally and did not perform well for standard $k=0.04$. The reason is that the Harris response function promotes corners and suppresses edges. The best results were obtained by employing $k=0$. This is a simpler definition for distinctiveness, but still it is not assessing adequately both the magnitude and the relative scale of $\lambda_1$ and $\lambda_2$. 
 
For the purpose of building appropriate distinctiveness maps, we designed a function that gives a high response to pixels that look like corners, lower to pixels that are parts of edges and, finally, a zero response to uniform areas. Furthermore, the distinctiveness is measured relatively within each image to ensure that all available information is exploited. In that direction, we first define $d$ to be the log of the magnitude of the vector $(\lambda_1, \lambda_2)$, i.e.,  
$
d = \log\left(\sqrt{\lambda_1^2 + \lambda_2^2}\right). 
$
The log function is used as a scaling operator. Subsequently, we compute the median $m_d$ of values $d$ over the whole image. Then, we define
\begin{equation}
\label{eq:conf_sig1}
d_s = \frac{1}{1 + e^{-(d-m_d)}}.
\end{equation}
$d_s$ is a sigmoid function that maps its input to the range $(0 .. 1)$. A $0.5$ response results for values $d$ equal to the median $m_d$ in the image. 
Similarly, we define $a$ as 
$    a = \arctan\left(\lambda_1,\lambda_2\right)$.
$a$ measures the difference between $\lambda_1$ and $\lambda_2$. Higher values for $a$ are indicative of a corner rather than an edge point. Given the median $m_a$ of values $a$ over the image, we define the function $a_s$ as:
\begin{equation}
\label{eq:conf_sig2}
a_s = \frac{1}{1 + e^{-(a-m_a)}}.
\end{equation}
By subtracting $m_d$, $m_a$ from $d$ and $a$ in Eq.~\ref{eq:conf_sig1} and Eq.~\ref{eq:conf_sig2}, respectively, we achieve relative $d_s$ and $a_s$ responses over the images. The product $d_s \cdot a_s$ represents the relative distinctiveness $c$ of a point:
\begin{equation}
\label{eq:conf_rhres}
c = \begin{cases}
                d_s \cdot a_s &\text{if }d_s \cdot a_s >w_T   \\
               0  &\text{otherwise}.  
   \end{cases}
\end{equation}
Points for which $d_s \cdot a_s \leq w_T$ are set to zero signifying that points that are less distinctive than a threshold are not considered at all. The threshold $w_T$ was determined experimentally as detailed in Section~\ref{sec:results_lbjective}. 
By measuring the value of $c$ for each pixel in the left and the right images of the stereo pair we obtain two distinctiveness maps, $C_l$ and $C_r$. 
 
\subsection{Rating a model hypothesis}
\label{sec:objective}
We denote with $I_l$, $I_r$ the two images of the stereo pair and with  $C_l$, $C_r$ the corresponding distinctiveness maps as computed in Section~\ref{sec:conf}. We consider a hypothesis $H$ about the configuration of the modelled scene (position, orientation, possible articulation) of all modelled and considered hands and objects to be tracked (see Section~\ref{sec:model}). Using the intrinsic and extrinsic parameters of the stereo, we can estimate the projection $p_l$ and $p_r$ of a 3D point $P_H= (X, Y, Z)$ of $H$ in each of the two views. Then, we define the color consistency  $s(p_l,p_r)$ of points $p_l$, $p_r$ as:
\begin{equation}
\label{eq:score}
s(p_l,p_r) = \min\{C_l(p_l),C_r(p_r)\} \cdot e^{-\beta \cdot ||I_l(p_l)-I_r(p_r)||}.
\end{equation} 
Intuitively, this color consistency measure considers the minimum of the distinctivenesses of the two points, scaled by a function that takes its maximum value when corresponding colors are identical and drops to zero with increasing color difference.  
In Eq.~\ref{eq:score}, $\beta$ is a scale parameter that controls the steepness of the exponential. The value of $\beta$ was determined experimentally (see Section~\ref{sec:results_lbjective}).
The total color consistency $S_H(I_l,I_r)$ of $I_l$, $I_r$ is then defined as 
\begin{equation}
\label{eq:tscore}
S_H(I_l,I_r) = \sum \limits_{p_l \in R_l^H, p_r \in R_r^H} s(p_l,p_r). 
\end{equation} 
In Eq.~\ref{eq:tscore}, $R_l^H$, and $R_r^H$ are the sets of points corresponding to the visible surface of $H$ in each view.

Care must be taken to exclude from consideration model points that are visible in one view but occluded in the other. To this end, we render the 3D model $H$ in both views. For each model point that is {\sl actually visible} in one view, we consider its projection in the other view. If the same physical point is visible (no occlusion), then we expect a rendered 3D point at exactly the same position (ideally) or within a short range $r$ (in practice). Otherwise, the point is excluded from consideration. We set $r=3$mm in all experiments.  
 
For a specific time instant, estimating the scene state $H$ amounts to estimating the optimal hypothesis $H^*$ that maximizes the objective function of Eq.~\ref{eq:tscore}, i.e.:
\begin{equation}
\label{eq:solution}
H^* = \arg \max_H \{ S_H(I_l,I_r) \}. 
\end{equation} 
 
\subsection{Stochastic optimization}
\label{sec:pso}
The optimization (maximization) problem defined in Eq.~\ref{eq:solution} is solved based on Particle Swarm Optimization (PSO)~\cite{james2001swarm} which is a stochastic, evolutionary optimization method. It has been demonstrated that PSO is a very effective and efficient method for solving vision optimization problems such as head pose estimation~\cite{padeleris2012head}, hand articulation tracking~\cite{oikonomidis2012tracking} and others. PSO achieves optimization based on the collective behavior of a set of particles (candidate solutions) that evolve in runs called generations. The rules that govern the behavior of particles emulate ``social interaction''. A population of particles is a set of points in the parameter space of the objective function to be optimized. PSO has a number of attractive properties. For example, depends on very few parameters, does not require differentiation of the objective function and converges with a relatively small computational budget~\cite{Angeline.PSO}. 
 
Every particle holds its current position (current candidate solution, set of parameters) in a vector $x_{t}$ and its current velocity in a vector $v_{t}$. Each particle $i$ keeps in vector $p_{i}$ the position at which it achieved, up to the current generation $t$, the best value of the objective function. The swarm as a whole, stores the best position $p_{g}$ across all particles of the swarm. All particles are aware of the global optimum $p_{g}$. The velocity and position update equations in every generation $t$ are
$ v_{t} = K(v_{t-1} + c_{1}r_{1}(p_{i} - x_{t-1}) + c_{2}r_{2}(p_{g} - x_{t-1}))$
and 
$    x_{t} = x_{t-1} + v_{t}$,
where $K$ is a constant \emph{constriction factor}~\cite{Clerc2002}, 
$c_{1}$ is called the \emph{cognitive component}, $c_{2}$ is termed the \emph{social component} and $r_{1}, r_{2}$ are random samples of a uniform distribution in the range $[0..1]$. Finally, $c_{1} + c_{2} > 4$ must
hold~\cite{Clerc2002}. As suggested in~\cite{Clerc2002} we set $c_{1} = 2.8$, $c_{2} = 1.3$ and 
$ %
K = \frac{2}{\left|2-\psi -\sqrt{\psi ^2-4\psi }\right|},
$ %
with $\psi = c_1 + c_2$.
 
In our problem formulation, the solution space has $N$ dimensions where $N$ is the sum of the parameters encoding the degrees of freedom of all the observed/tracked objects. Specifically, $N = 27$ for tracking a single hand, $N = 27+7=34$ for a hand interacting with a rigid object and $N = 27+27=54$ for two hands. Particles are initialized with a normal distribution around the center of the search range with their velocities set to zero. Each dimension of the multidimensional parameter space is bounded in some range. During the position update, a velocity component may force a particle to move to a point outside the bounded search space. Such a component is truncated and the particle does not move beyond the boundary of the corresponding dimension. Since the object(s) motion needs to be continuously tracked in a sequence instead of being estimated in a single frame, temporal continuity is exploited. More specifically, the solution over frame $t$ is used to restrict the search space for the initial population at frame $t+1$. In related experiments, the search range (or the space in which particle positions are initialized) extend $\pm 40\, mm$ (for positional parameters) and $\pm 10^\circ$ (for rotational parameters) around their estimated values in the previous frame.
 
\begin{figure}[t]
\begin{center}
\includegraphics[width=0.48\textwidth]{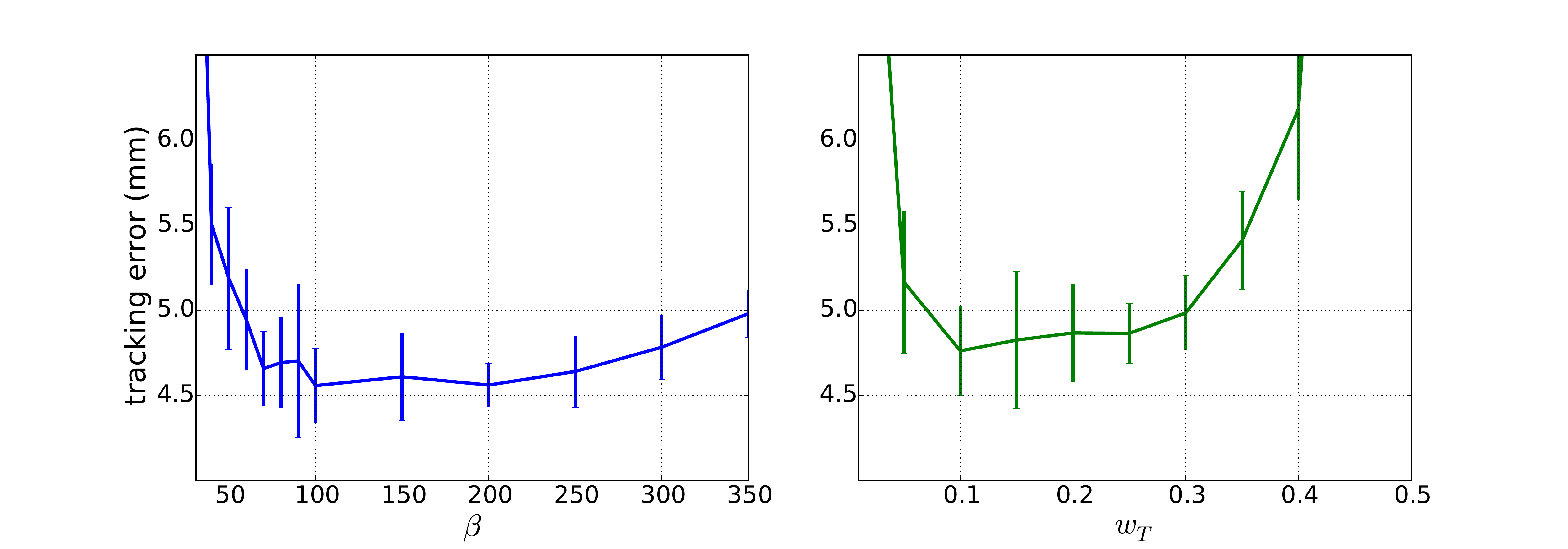} 
\caption{\label{fig:params} Tracking error for different values of the parameters $\beta$ (left) and $w_T$ (right). See text for details.}
 \end{center}
\end{figure}

\begin{figure*}[t]
\begin{center}
\includegraphics[width=0.31\textwidth]{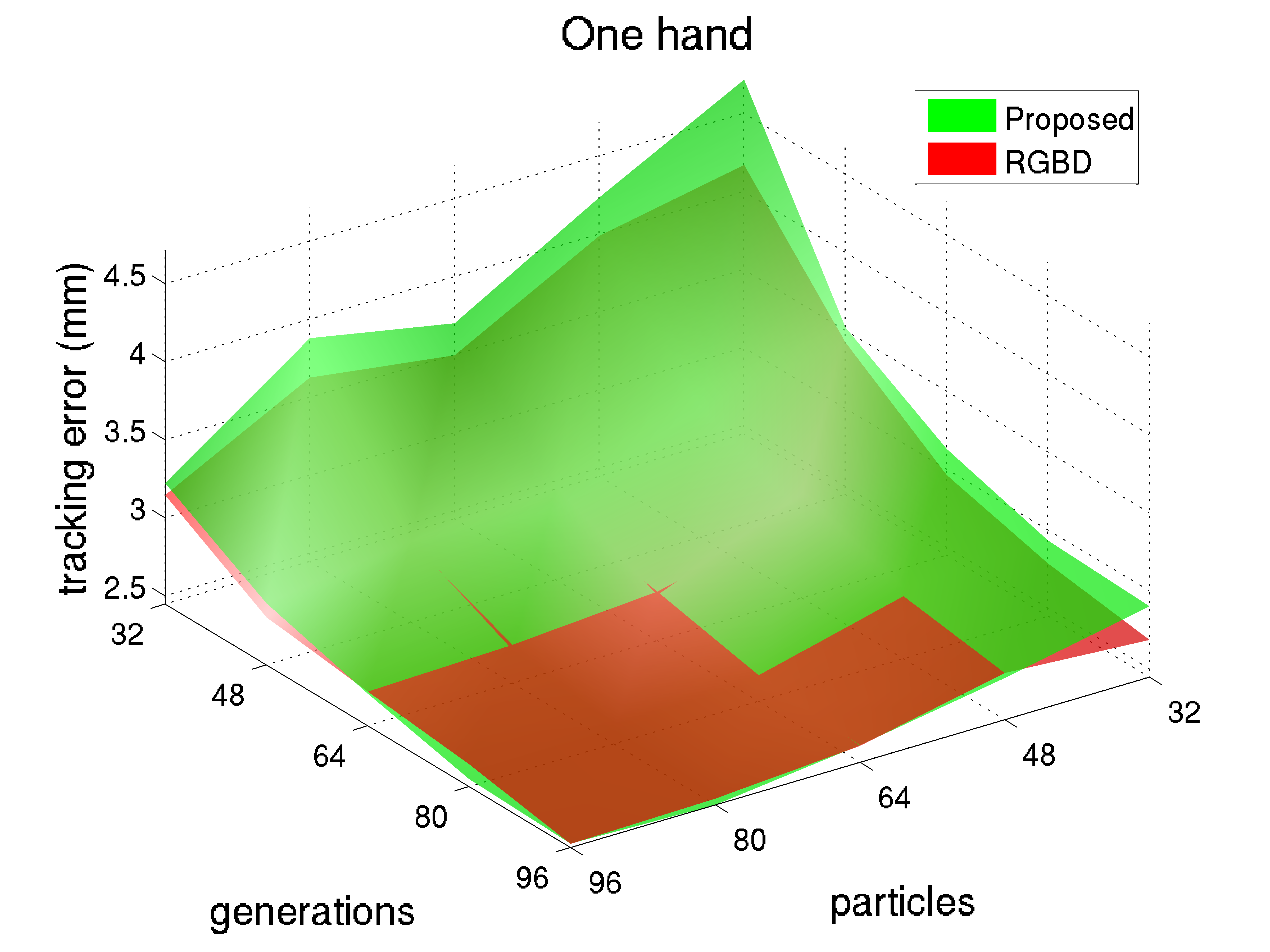} 
\includegraphics[width=0.31\textwidth]{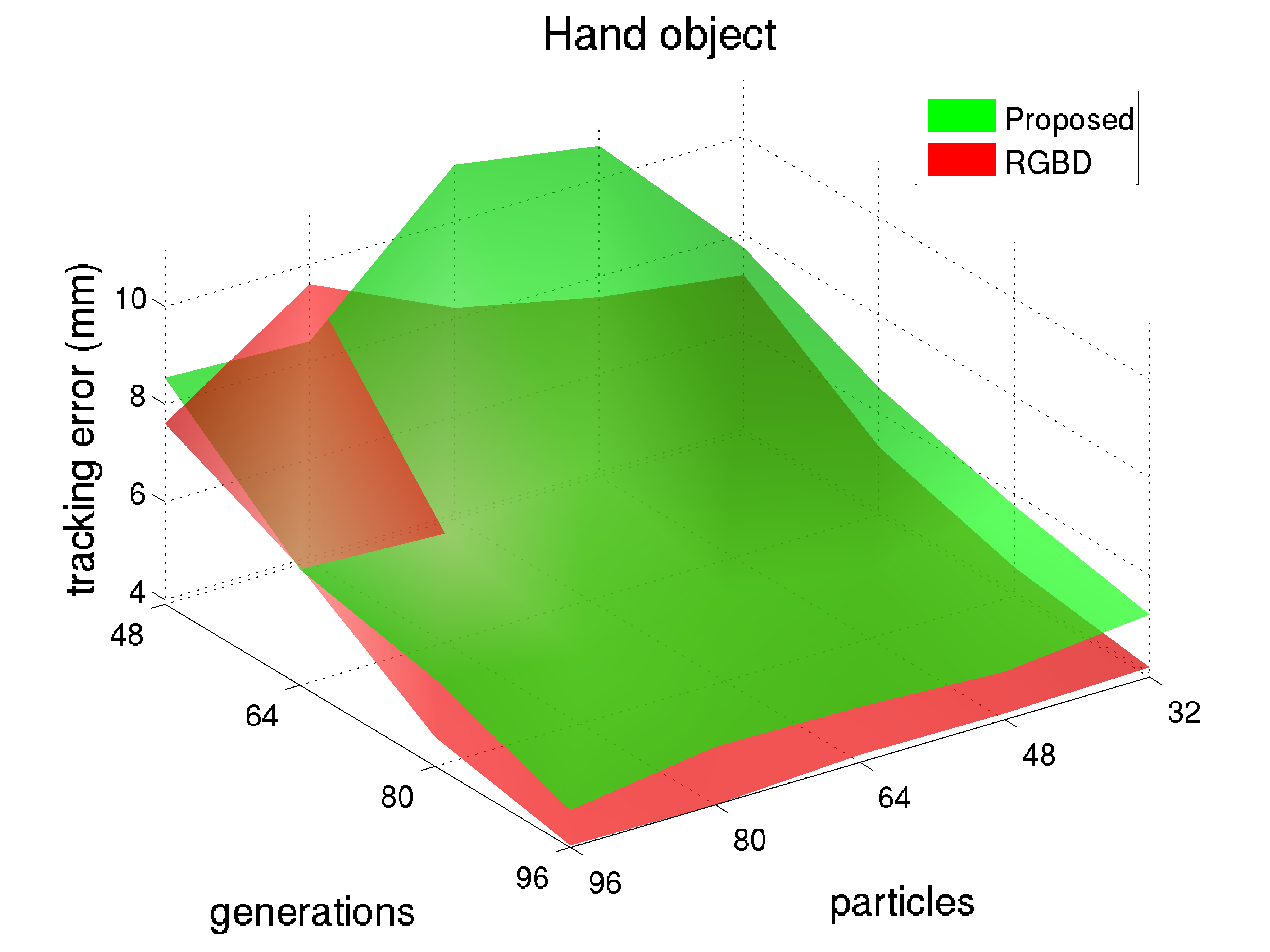} 
\includegraphics[width=0.31\textwidth]{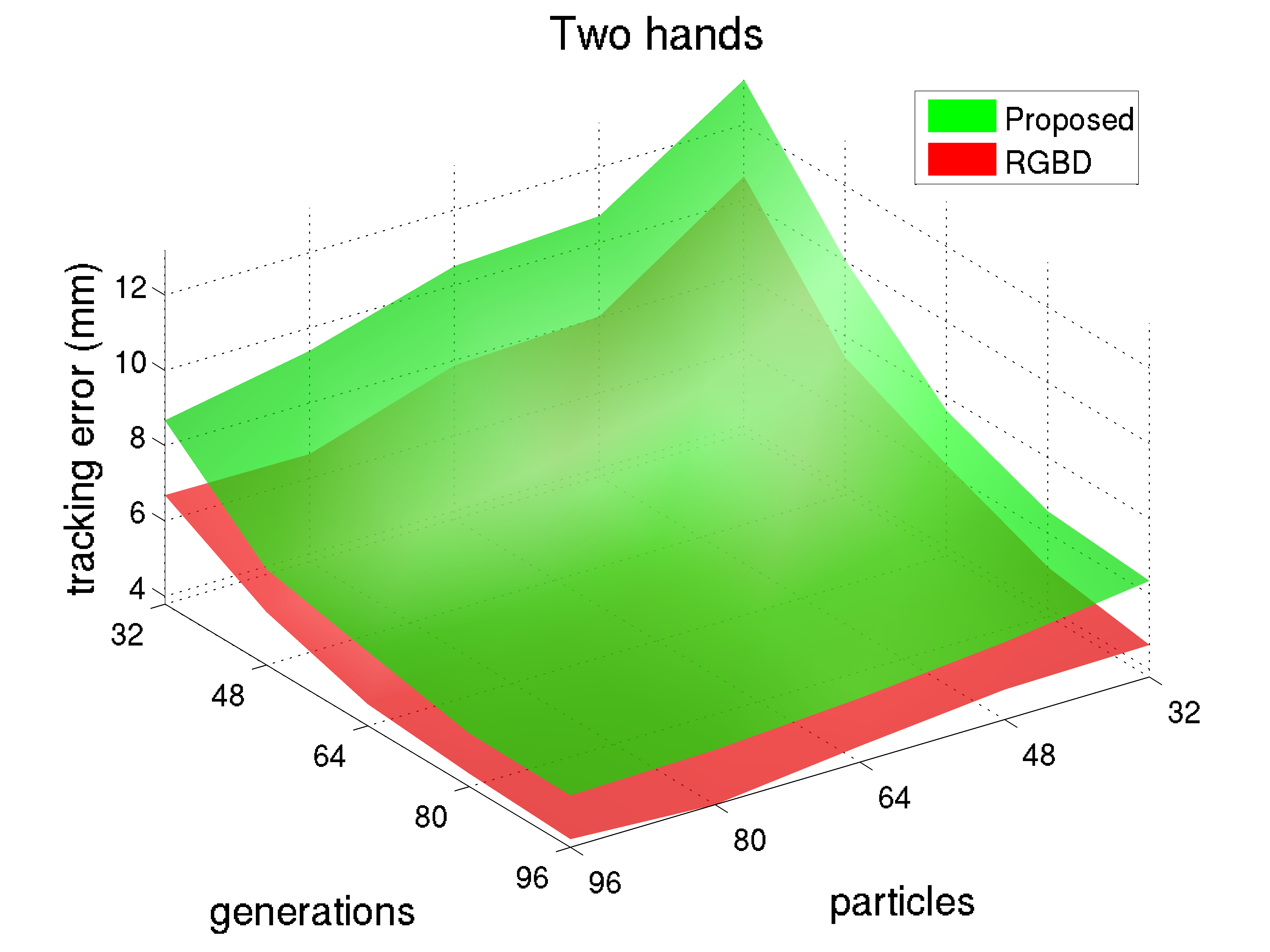} 
\includegraphics[width=0.31\textwidth]{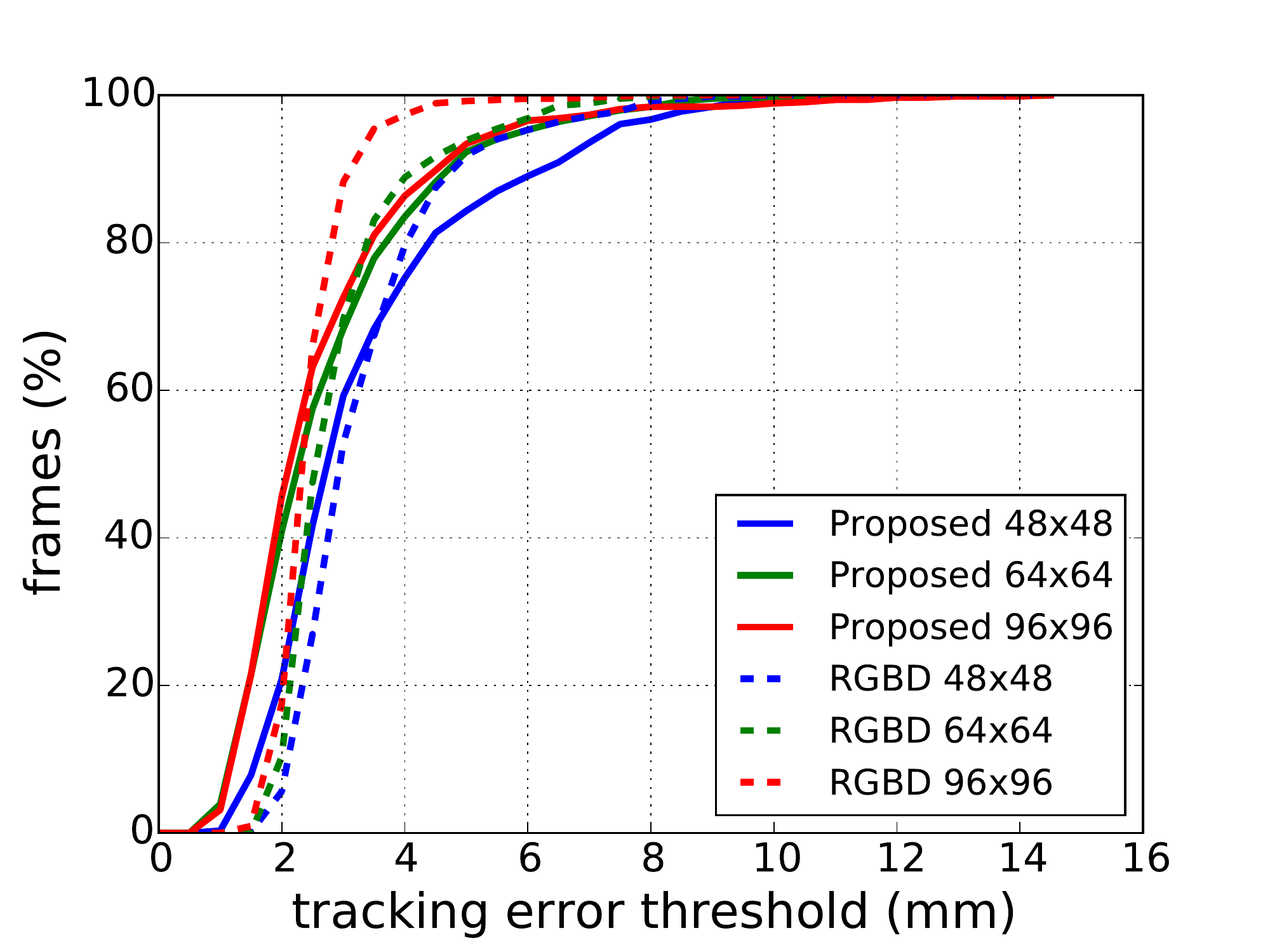} 
\includegraphics[width=0.31\textwidth]{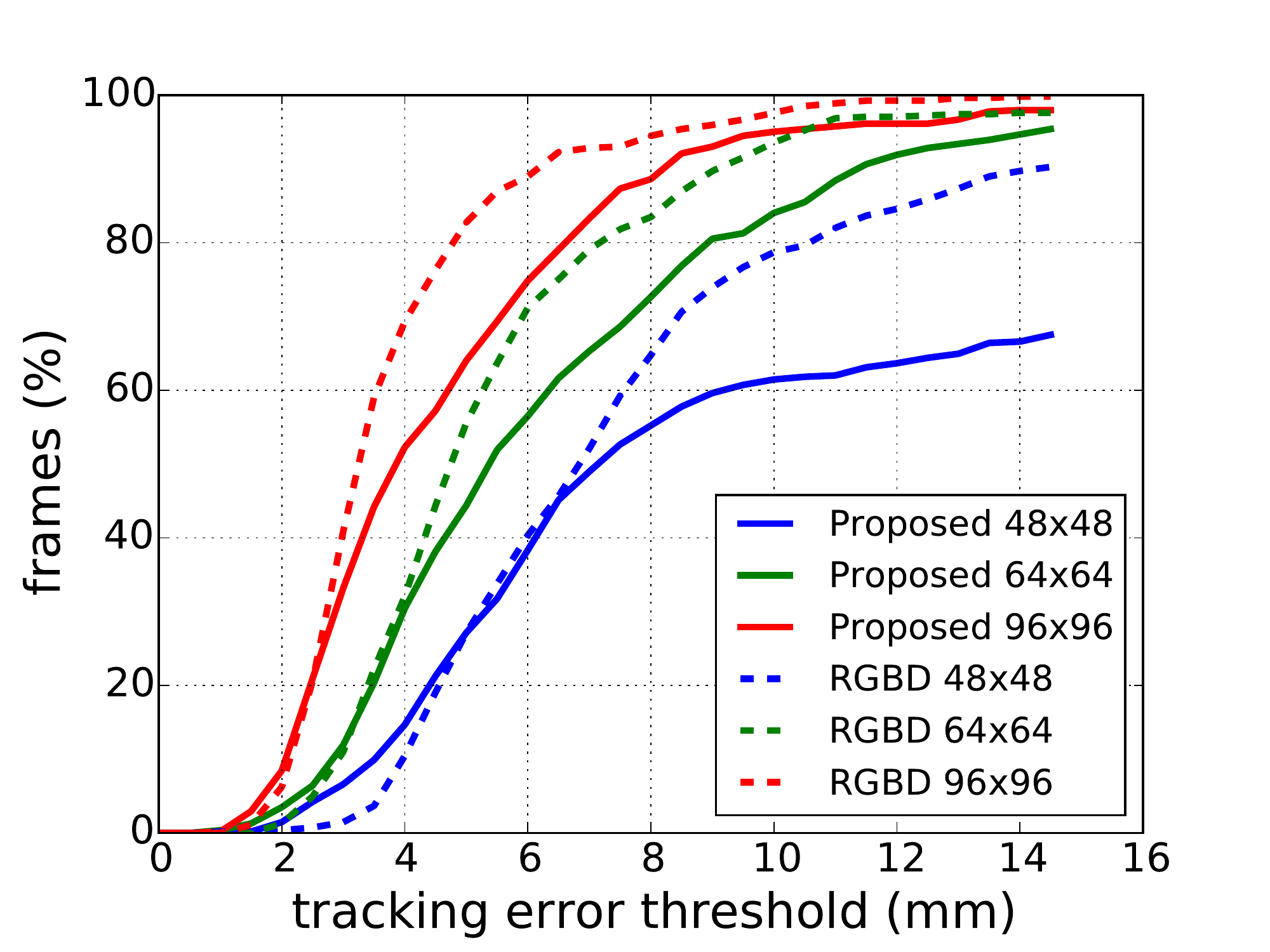} 
\includegraphics[width=0.31\textwidth]{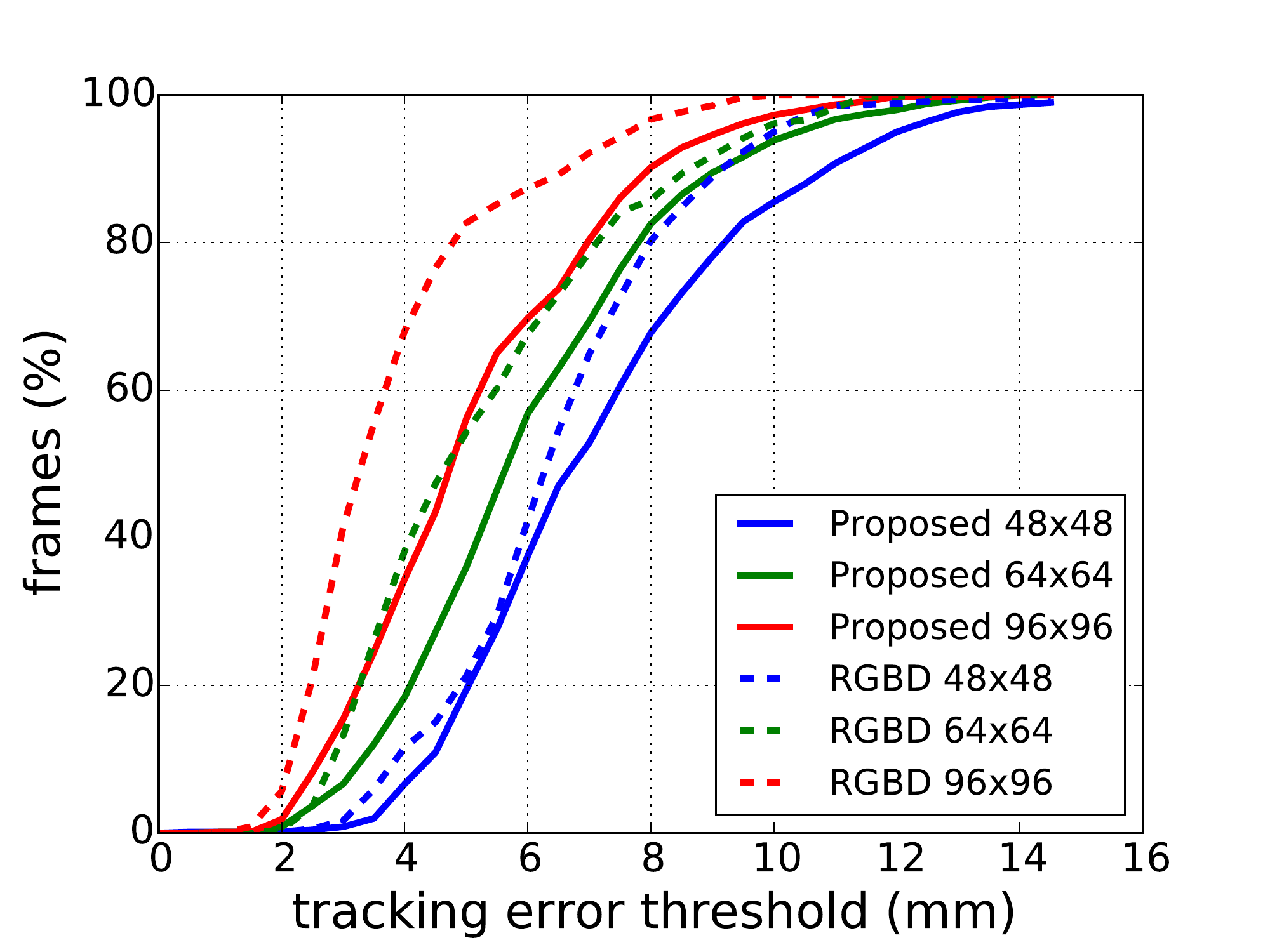} 
\caption{\label{fig:quant} Quantitative evaluation of the proposed method: Columns (left to right), single hand, hand-object, two hands. Top row: 3D tracking error as a function of the PSO particles and the generations for the baseline methods and the proposed one. Bottom row: percentage of successfully tracked frames for the baseline methods (dashed) and the proposed one with different budgets as a function of the error tolerance (see text for details).}
 \end{center}
\end{figure*}

\subsection{Implementation and performance issues}
\label{sec:implementation}
In order to achieve a real time frame rate, the proposed pipeline was implemented using CUDA on an NVIDIA GPU. 
In order to lower the computational requirements, the input images are segmented around the last known solution. The bounding box of the tracked objects is computed for each frame by rendering a synthetic view of the scene in full resolution as it appeared in the previous frame and then segmenting an area around the objects. The distinctiveness maps are computed only for the segmented images. 

PSO is inherently parallelizable since the particles are only synchronized at the end of each generation. In our implementation we exploit this feature by evaluating the objective function for each particle in parallel, using CUDA. For each generation all hypothesized scene configurations are rendered using OpenGL.
Our reference implementation can achieve near real time performance ($15-20$fps) for single hand tracking running with a budget of $32$ particles and $32$ generations on a computer equipped with an Intel i7 950 @ 3.07GHz CPU, 12GB of RAM and an NVidia GTX970 GPU. Although the rendering of each generation can be performed in a single OpenGL drawing call, the reference implementation renders once for each camera view. Further optimizations in the pipeline implementation could increase the performance significantly on the same hardware.

\section{Experimental evaluation}
\label{sec:results}
While in recent years some datasets with ground truth for tracking articulated hand motions have been made available, they use RGBD sensors and not stereo. In order to evaluate the proposed method and provide a fair comparison to other methods, the standard protocols used in the relevant literature~\cite{OikonomidisKyriazisArgyros2011a, oikonomidis2010markerless, kyriazis2014scalable} were followed. Synthetic stereo and RGBD datasets were created using the tracking results of baseline RGBD methods on real world sequences. The scenarios in the datasets consist of articulations of (a)~a single hand, (b)~a hand interacting with an object and (c)~two interacting hands. To evaluate quantitatively the performance of our approach, we directly compare it with state of the art model-based methods that use RGBD input. For the experiments, we used our implementation of~\cite{oikonomidis2011efficient} for single hand tracking and our implementation of~\cite{oikonomidis2012tracking} for two hands tracking. A variant of~\cite{oikonomidis2012tracking} was also used to track a hand interacting with a rigid object. In our implementations, the above methods were adapted so as to operate with skinned hand models instead of hand models consisting of collections of geometric solids. This reduces the tracking error of the baseline approaches over the original methods reported in \cite{oikonomidis2011efficient,oikonomidis2012tracking}. 
In the qualitative experiments, we considered multiple sequences covering several indoors and outdoors scenarios.

\subsection{Datasets}
{\noindent \bf Synthetic data:} 
For the single and two hands tracking scenarios we used\footnote{Became available to us after contacting the authors of~\cite{oikonomidis2011efficient,oikonomidis2012tracking}.} the synthetic sequences and ground truth presented in \cite{oikonomidis2011efficient,oikonomidis2012tracking}.  
For the hand-object tracking scenario we captured an RGBD sequence of a human manipulating a spray bottle. The spray bottle model was created with a laser scanner which provides millimetre accuracy. The hand-object sequence was tracked with the variant of ~\cite{oikonomidis2012tracking} in order to obtain a ground truth.
Subsequently, the ground truth from the three sequences was used to render (a)  {\sl synthetic stereo} datasets to be used by our method and (b) {\sl synthetic RGBD} to be used by the baseline RGBD methods we compare against. 
For the synthetic RGB images, the texture information of the hands and object were used during rendering. The rendered models were illuminated using an ambient light source. Real world images of indoor environments were captured with the ZED sensor and used as the background of the models for the synthetic stereo. 
 
The single hand ({\bf SH}) sequence consists of 638 frames, the hand-object sequence ({\bf HO}) of 545 frames and the two-hands sequence ({\bf TH}) of 705 frames. All sequences cover challenging articulations of hands as well as hand-object and hand-hand interactions. To the best of our knowledge, no other datasets with ground truth exist for tracking hand-object and hand-hand interactions with a stereo pair. We plan to make all three sequences publicly available.
 
{\noindent \bf Real world sequences:} In order to support the qualitative evaluation of the proposed method on real data, stereo sequences were captured using the ZED stereo pair~\cite{zedcamera} both indoors and outdoors. Three different groups of sequences were recorded, for the problems of tracking a single hand, hand-object interaction and two hands. The real world sequences were recorded at different combinations of resolutions and frame rates (1080p, 720p, 30fps and 60fps) both indoors and outdoors. The sequences show a user performing various hand motions, manipulating an object in front of the camera and performing bi-manual hand gestures. The hand models used (see Section~\ref{sec:model}) do not exactly match the user's hands (e.g., differ in size and finger length). For the hand-object real-world sequences the object used was a pencil-box. The box was modelled as a cuboid that is just an approximation to the actual object's 3D shape. It is demonstrated that the proposed method is robust to these inconsistencies between the scene models and the actual observations. 
 
\subsection{Performance metrics}
We use standard metrics to assess the tracking error of the evaluated methods. For a hand, the tracking error is computed as the mean distance of the corresponding hand joints from the ground truth. For an object model, we use $3$ anchor points on the object's model. The error of the object pose is the mean distance of the tracked anchor points from the corresponding ground truth. The tracking error for a frame is the average tracking error of all the objects it involves. Finally, the tracking error of a sequence is the average per-frame tracking error.
 
\subsection{Results}
{\noindent \bf Deciding internal parameters:}
\label{sec:results_lbjective}
As presented in Section~\ref{sec:method}, the proposed method entails the setting of (a) the distinctiveness threshold $w_T$ that controls the image points that contribute to the objective function~(Eq.~\ref{eq:conf_rhres}) and (b) $\beta$, which gauges the steepness of the color similarity curve (Eq.~\ref{eq:score}).  Using the synthetic dataset for the single hand, we investigated the effect of different parameter values on the accuracy of the method. Figure~\ref{fig:params} (left), shows the tracking error for a range of $\beta$ values ($30 \leq \beta \leq 350$) and  $w_T=0.1$. The right plot in this figure shows the tracking error for  $0.0 \leq w_T \leq 0.45$ and $\beta=100$. The tests are performed with a relatively restricted PSO budget (32 particles and 32 generations) where a fine tuned objective function is needed to achieve good results and, thus, performance differences are more prominent. The plots show the mean tracking error over multiple runs as well  as the standard deviation. It can be verified that good performance is attained for a wide range of values for both parameters, indicating that the method is not sensitive to their exact setting. 
The values $w_T=0.1$ and $\beta=100$ were used in all experiments.

\begin{figure}[t]
\begin{center}
\includegraphics[width=0.32\columnwidth]{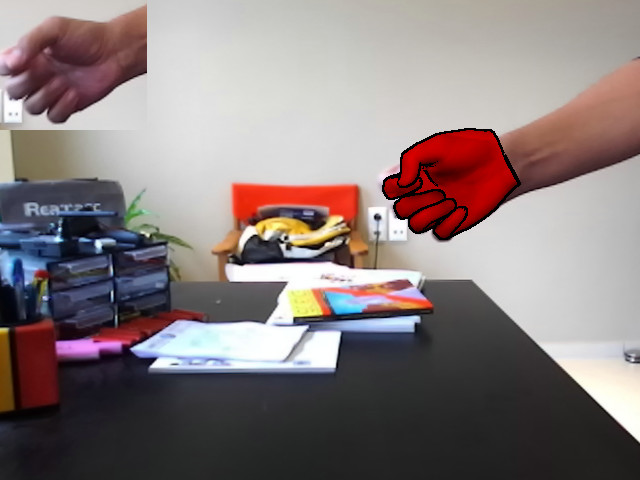}
\includegraphics[width=0.32\columnwidth]{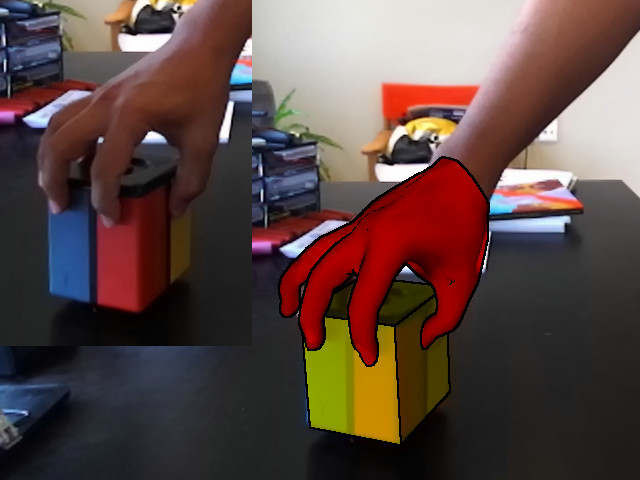}
\includegraphics[width=0.32\columnwidth]{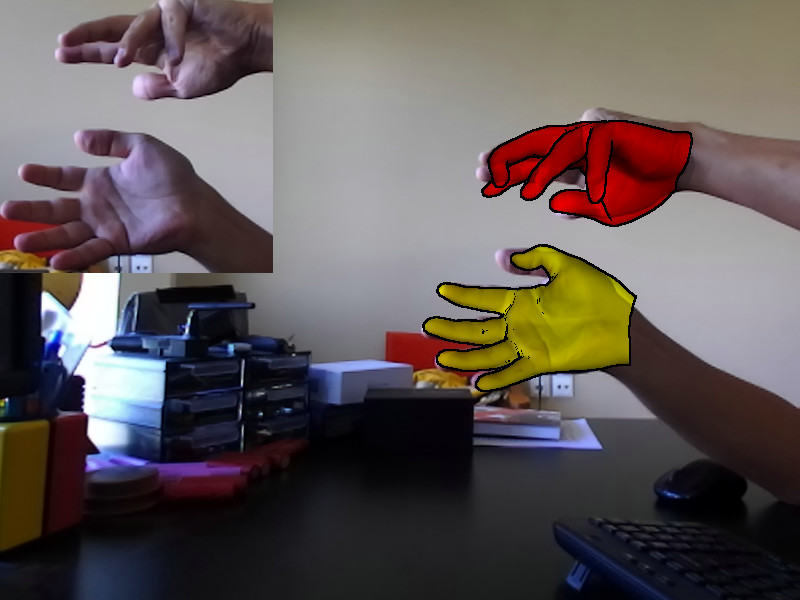}
 
\includegraphics[width=0.32\columnwidth]{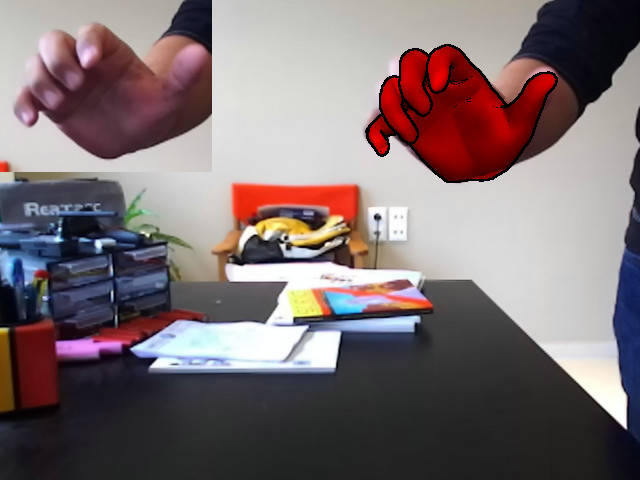}
\includegraphics[width=0.32\columnwidth]{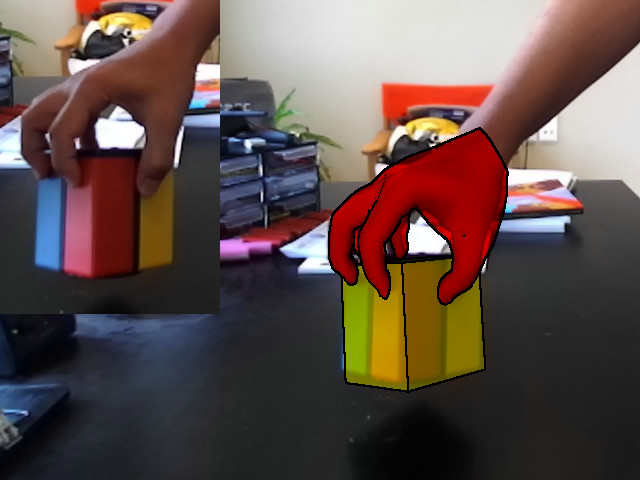}
\includegraphics[width=0.32\columnwidth]{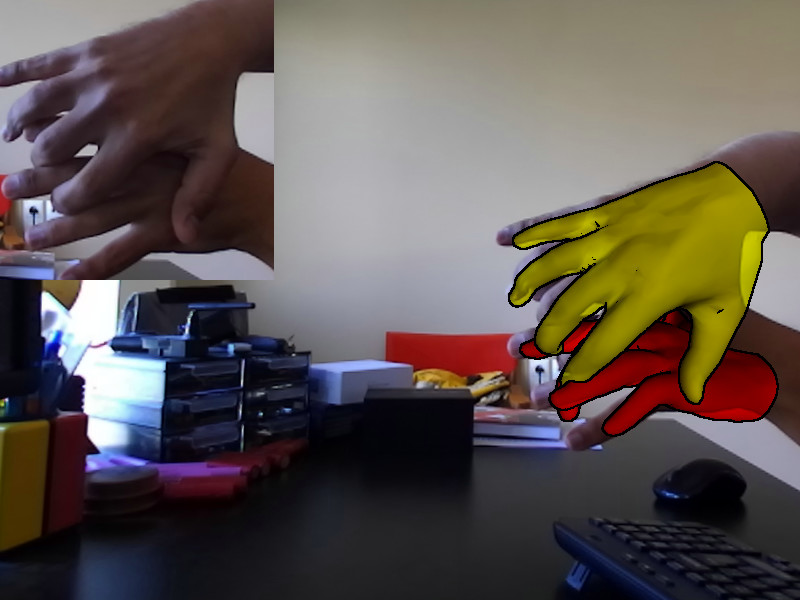}
 
\includegraphics[width=0.32\columnwidth]{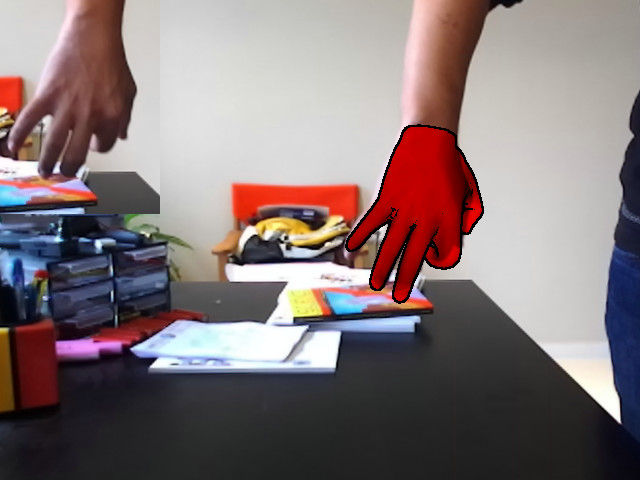}
\includegraphics[width=0.32\columnwidth]{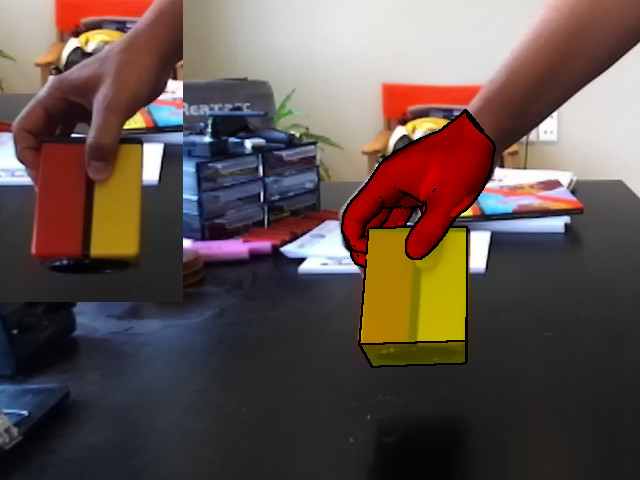}
\includegraphics[width=0.32\columnwidth]{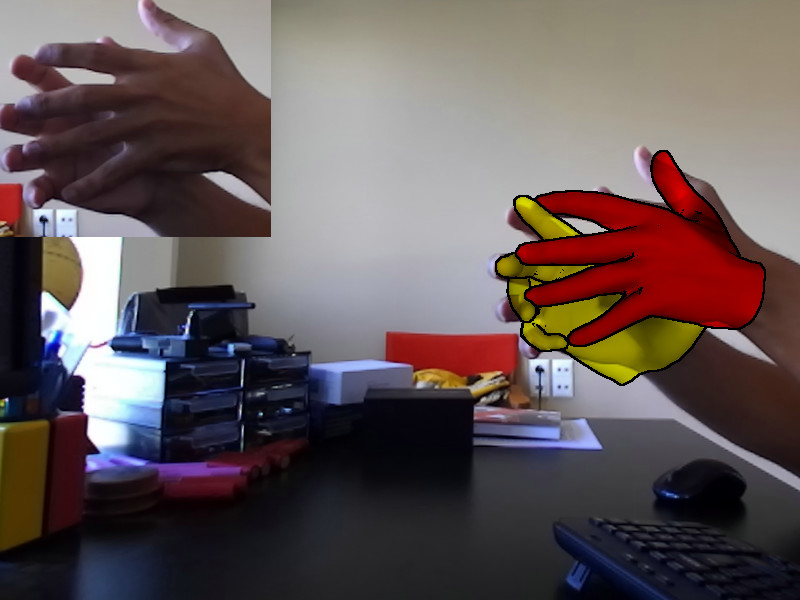}
                         
\includegraphics[width=0.32\columnwidth]{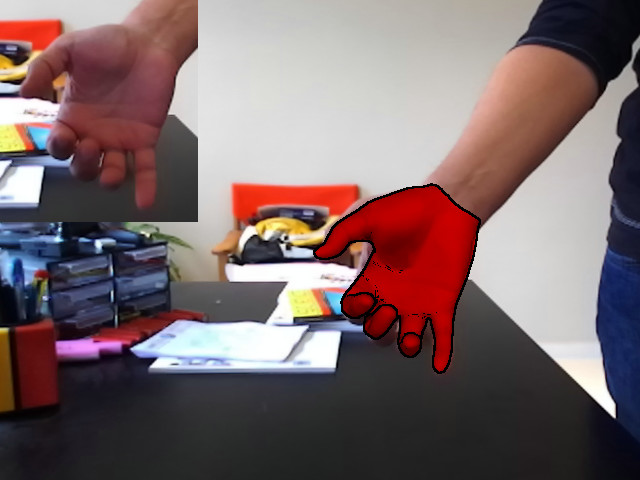}
\includegraphics[width=0.32\columnwidth]{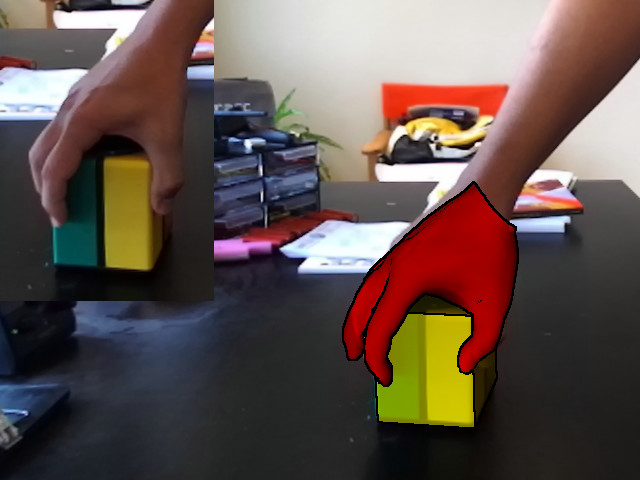}
\includegraphics[width=0.32\columnwidth]{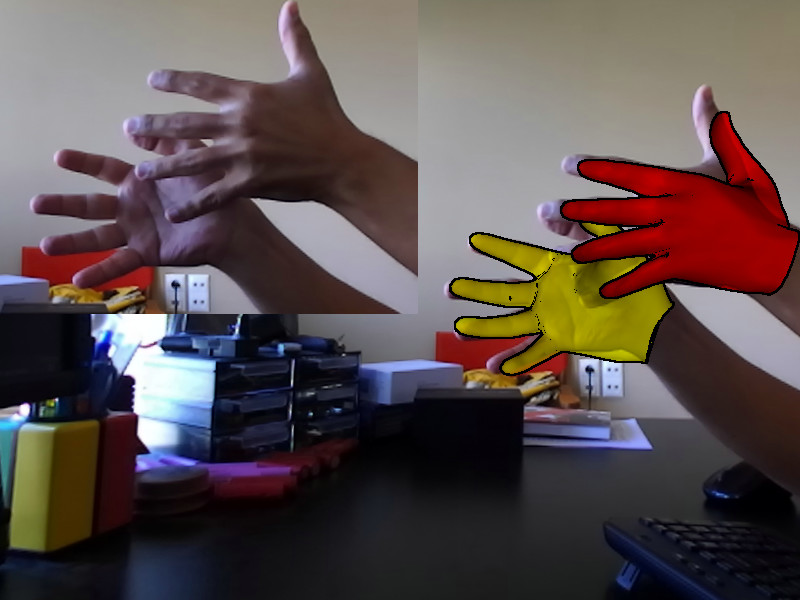}
                         
\includegraphics[width=0.32\columnwidth]{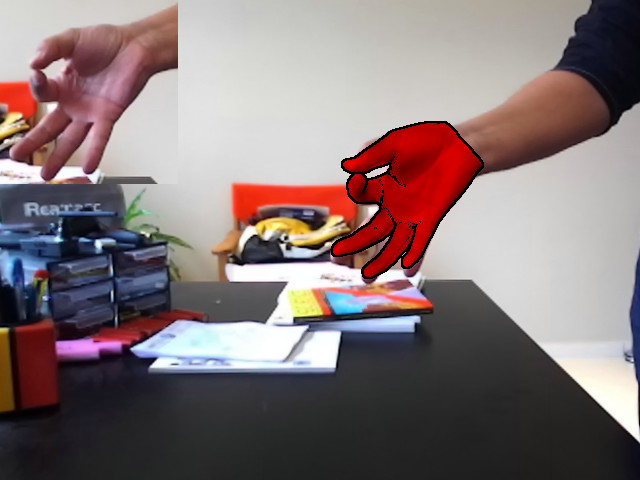} 
\includegraphics[width=0.32\columnwidth]{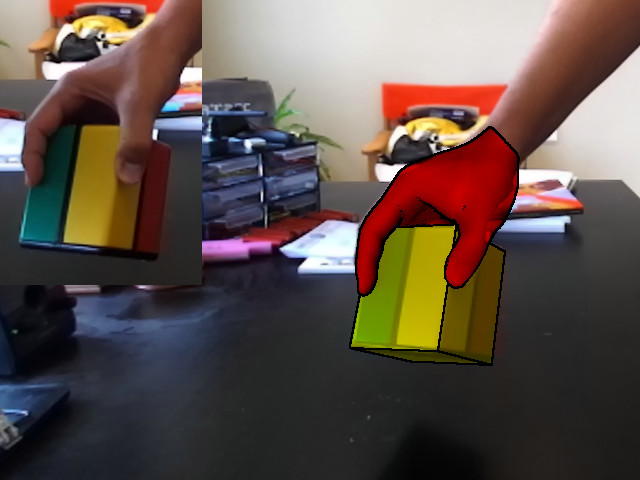}
\includegraphics[width=0.32\columnwidth]{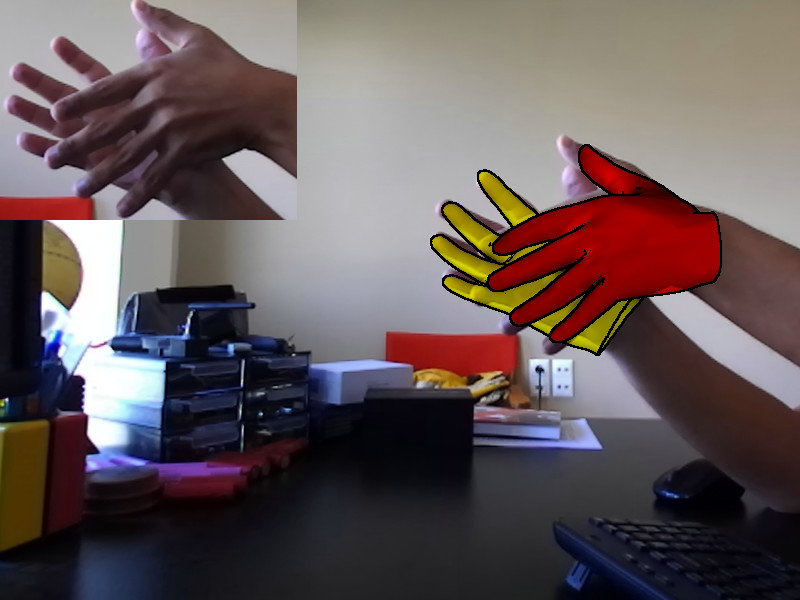}
 
\includegraphics[width=0.32\columnwidth]{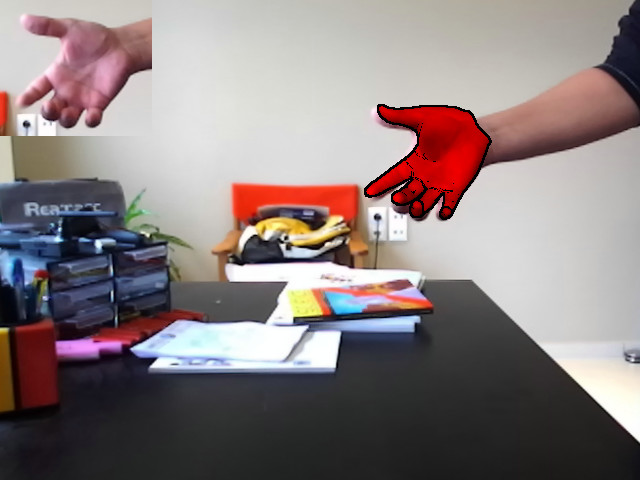}
\includegraphics[width=0.32\columnwidth]{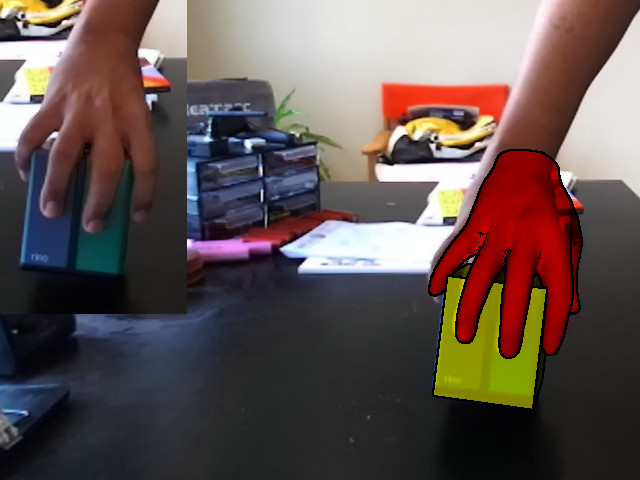}
\includegraphics[width=0.32\columnwidth]{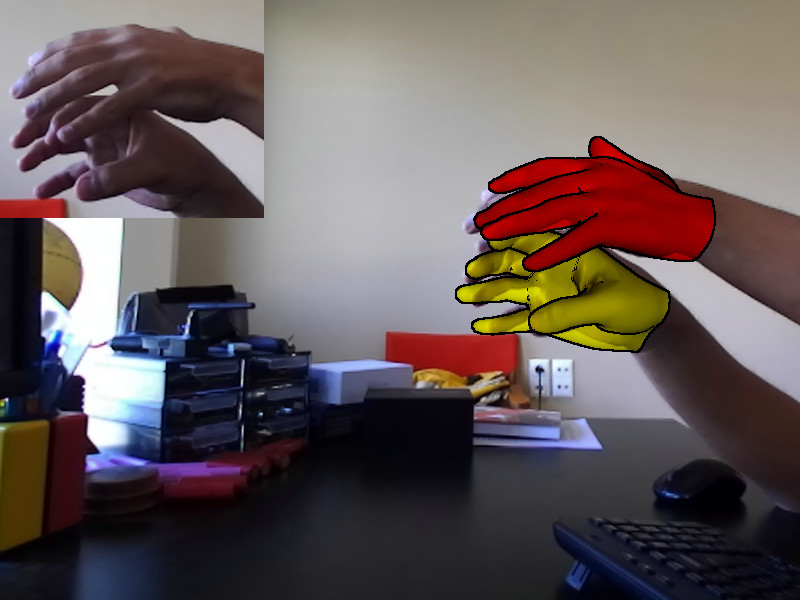}
 
\caption{\label{fig:qualitative} Qualitative evaluation. Results on real world sequences. Tracking of a single hand sequence (left), of a hand interacting with a pencil box (middle) and of two interacting hands (right).}
 \end{center}
\end{figure}

\begin{figure}[t]
\begin{center}
\includegraphics[width=0.24\columnwidth]{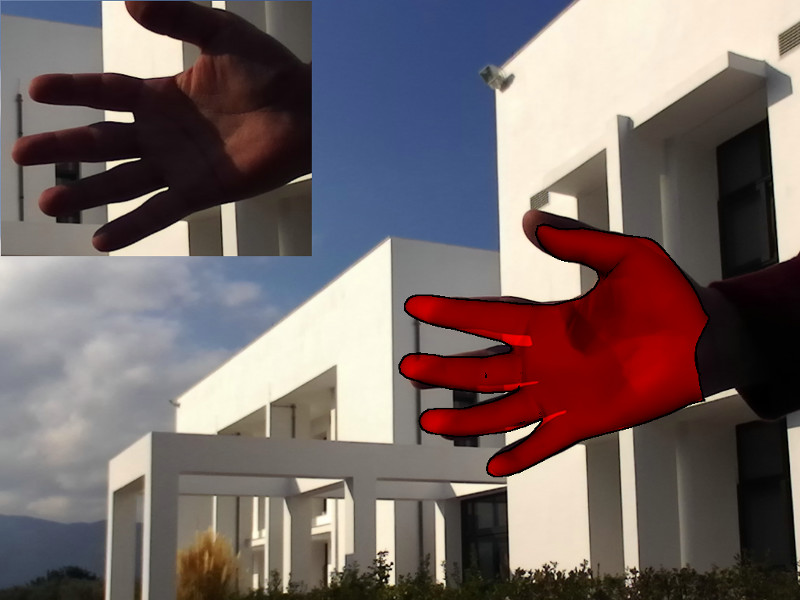}
\includegraphics[width=0.24\columnwidth]{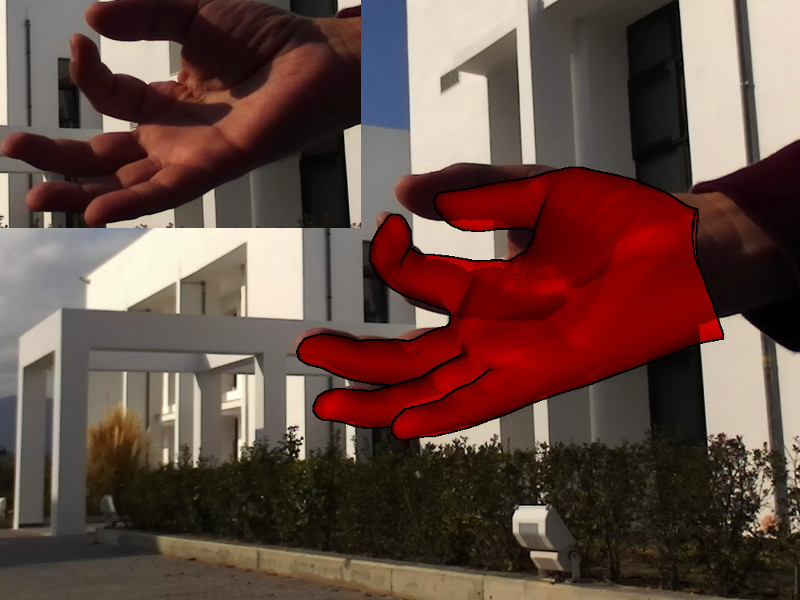}
\includegraphics[width=0.24\columnwidth]{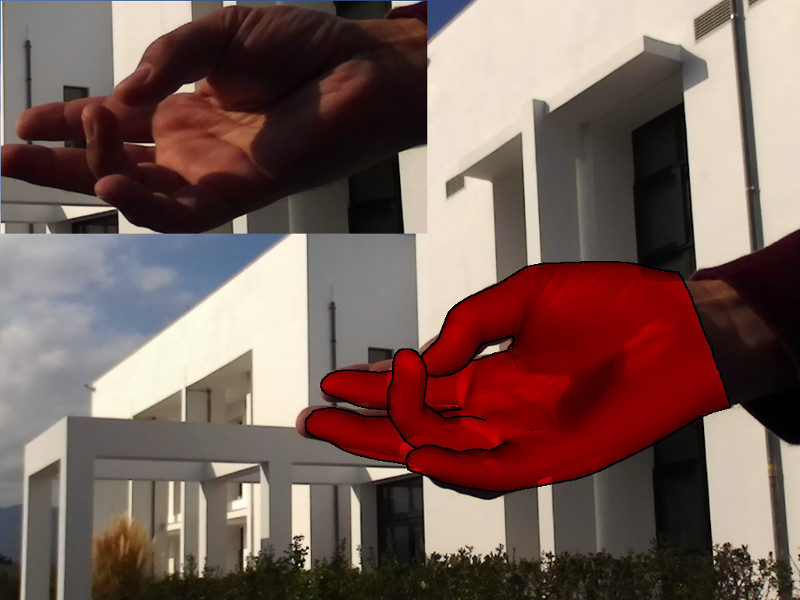} 
\includegraphics[width=0.24\columnwidth]{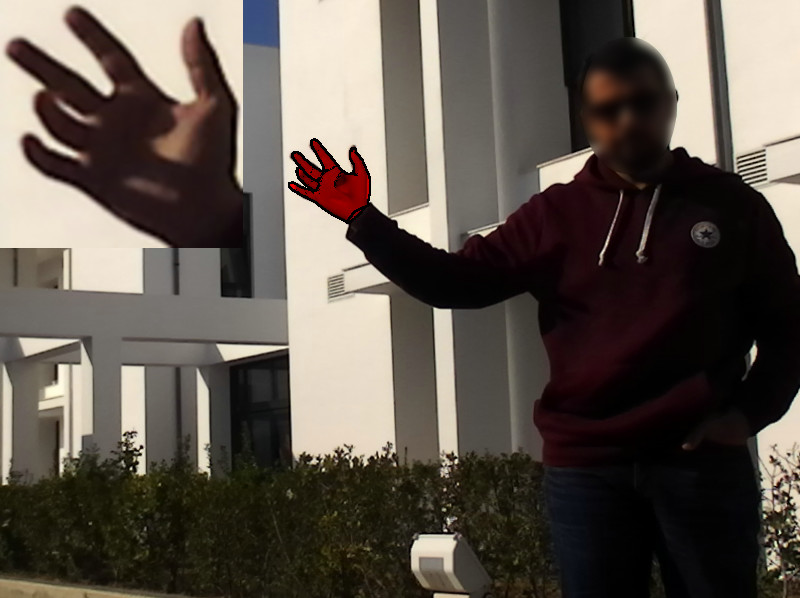}
\caption{\label{fig:qualitative_outdoors} Tracking of a single hand on an outdoors sequence. The tracking result is superimposed on the image while the segmented hand is shown in the top left of each frame.}
 \end{center}
\end{figure}

\begin{figure}[t]
\begin{center}
\includegraphics[width=0.32\columnwidth]{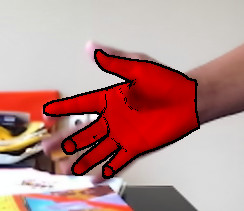}
\includegraphics[width=0.32\columnwidth]{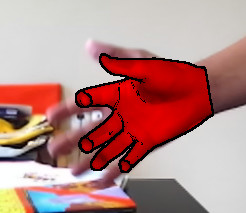}
\includegraphics[width=0.32\columnwidth]{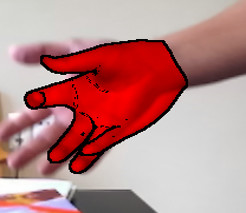}
 
\includegraphics[width=0.32\columnwidth]{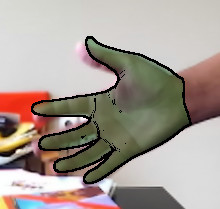}
\includegraphics[width=0.32\columnwidth]{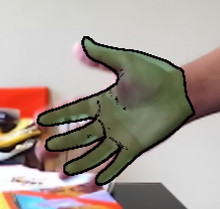}
\includegraphics[width=0.32\columnwidth]{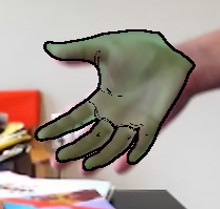}
\caption{\label{fig:zeddepth} Top row: applying the approach of~\cite{oikonomidis2011efficient} on the depth maps produced by the ZED camera fails immediately (tracking result in red). Bottom row: the proposed method tracks correctly the hand (tracking result in  green). From left to right: frames 1 (initialization), 2 and 15.
}
\end{center}
\end{figure}
 
{\noindent \bf Quantitative evaluation of tracking accuracy:}
\label{sec:quantitative}
Using the synthetic sequences we measure the tracking error of the proposed method as a function of different PSO budget configurations. In these the experiments, both methods (proposed and baseline) are initialized at the first frame of the sequence using the known ground truth for the frame (same initial position). 
Tracking failures were not re-initialized. Each plot on the top row of Figure~\ref{fig:quant} shows the tracking error of the proposed method and its RGBD competitor for the {\bf SH}, {\bf HO} and {\bf TH} datasets (left to right). For each data point we consider the mean over $10$ runs.
 
The proposed method performs similarly or better for the {\bf SH} dataset when compared to~\cite{oikonomidis2011efficient} despite the fact that the proposed method relies on much less (or implicit) information compared to~\cite{oikonomidis2011efficient}. For the problems of higher dimensionality the results indicate that RGBD solutions are more accurate but only by a small margin. 
 
The bottom row of Figure~\ref{fig:quant} shows the percentage of frames of a dataset for which the tracking error was below a certain threshold. There are six curves in each plot. The dashed curves correspond to the RGBD baseline methods. Each color corresponds to a different PSO budget (particles$\times$generations). It is interesting to note that when given a higher computational budget, the proposed method can achieve similar or better accuracy to the RGBD counterpart on the {\bf HO} and {\bf TH} datasets.
  
{\noindent \bf Assessing the effect of foreground detection:}
The RGBD methods use segmented depth as input. Typically, this is achieved either using skin color detection on the RGB input (less robust) or based on depth segmentation (more robust). In order to make a fair comparison between the proposed and the methods we are comparing against,  in the experiments reported in Figure~\ref{fig:quant}, we used the same masking method for all evaluated methods. However, in real life, our stereo-based approach does not have access to the more robust depth-based foreground segmentation. In that respect, we are interested in assessing the influence of foreground masking on the performance of the proposed method. We performed the same experiments, for different PSO budget configurations, with and without foreground masking.
Masking reduces tracking error but only marginally ($0.3$mm on average). Thus, foreground masking can be skipped without significantly affecting the tracking accuracy. 
 
{\noindent \bf Qualitative evaluation in real world sequences:}
\label{sec:qualitative}
Figure~\ref{fig:qualitative} shows representative frames with tracking results on the real world datasets. The computed pose is superimposed on the left RGB image of the stereo pair. Different objects are shown in different color (red and yellow). On the top left of each frame the original bounding box of the observation is shown. While depth-based methods fail outdoors due to ambient infrared light, our stereo based method does not suffer from this problem. In Figure~\ref{fig:qualitative_outdoors} we demonstrate results from an outdoors sequence tracking a single hand in short ($<0.5m$) and longer ($~3m$) distances. In all sequences, tracking is performed without any foreground segmentation. Complete results are provided in the supplementary material accompanying the paper.

{\noindent \bf Tracking by relying on depth from stereo:}
A straightforward idea for stereo-based tracking would be to first reconstruct 3D structure and then use the approach of~\cite{oikonomidis2011efficient,oikonomidis2012tracking} on the resulting depth maps. This approach is also explored in~\cite{zhang20163d}. We evaluated this alternative compared to our color consistency based approach that bypasses the problem of 3D reconstruction. Essentially, we fed the RGBD based approach with the stereo-based depth information provided by the ZED camera and made sure that the depth around the hand was segmented correctly. It turns out that this is a rather unreliable solution that fails very fast. This is due to the quality of the depth information which is either dense but very noisy, or reliable but sparse or smoothed out. Figure~\ref{fig:zeddepth} shows indicative results. Both methods are initialized at the same initial position on the first frame but they diverge quickly. The depth based method looses track after very few frames. The supplemental material provides further relevant evidence.

\section{Summary and conclusions}
\label{sec:conclusions}
We presented a novel approach for tracking hand interactions in various settings. The proposed method is the first that can cope accurately and efficiently with these tracking scenarios based on a conventional short baseline stereo. By employing a hypothesise-and-test framework, we cast tracking as an optimization problem that maximizes the color consistency of the tracked scene. Thus, we avoid the explicit computation of disparity maps which is particularly challenging for the relatively uniformly colored human hands. Our approach achieves similar, and in some cases, better tracking accuracy than state of the art methods that use specialized active sensors such as depth cameras, at a comparable computational performance. 
From a theoretical point of view, the significance of the proposed method is that it shows that accurate 3D structure information is not a prerequisite for 3D hand tracking and related problems. From a practical point of view, the significance of the developed method is also high, as it enables 3D hand tracking based on compact, conventional, widely deployed, passive vision systems that do not have the limitations of contemporary RGBD cameras (e.g. limited spatial resolution, constraints on distance of observation, sensitivity to infrared light). Moreover, although we focus on the challenging tracking scenarios involving hands, nothing prevents the applicability of the proposed algorithm to problems such as human skeleton tracking and 3D tracking of rigid objects.

 {\small
\bibliographystyle{ieee}
\bibliography{stereo}
}
 
\end{document}